\documentclass[journal]{IEEEtran}
\usepackage{enumerate}
\usepackage{amssymb}
\usepackage{amsmath}

\usepackage{graphicx}
\usepackage{subfigure}
\usepackage{bm}
\usepackage{cite}
\usepackage{multirow}
\usepackage{color}

\usepackage{makecell}
\usepackage[table,xcdraw]{xcolor}
\usepackage{booktabs} 
\usepackage{graphicx}
\usepackage[colorlinks=true, linkcolor=blue, urlcolor=blue, citecolor=blue]{hyperref}

\ifCLASSINFOpdf
\else
\fi
\begin{document}
%
% paper title
% Titles are generally capitalized except for words such as a, an, and, as,
% at, but, by, for, in, nor, of, on, or, the, to and up, which are usually
% not capitalized unless they are the first or last word of the title.
% Linebreaks \\ can be used within to get better formatting as desired.
% Do not put math or special symbols in the title.
\title{Weakly Supervised LiDAR Semantic Segmentation via Scatter Image Annotation}
%
%
% author names and IEEE memberships
% note positions of commas and nonbreaking spaces ( ~ ) LaTeX will not break
% a structure at a ~ so this keeps an author's name from being broken across
% two lines.
% use \thanks{} to gain access to the first footnote area
% a separate \thanks must be used for each paragraph as LaTeX2e's \thanks
% was not built to handle multiple paragraphs
%
\author{Yilong~Chen,
        Zongyi~Xu,
        Xiaoshui~Huang,
        Shanshan~Zhao,
        Xinqi~Jiang,
        Xinyu~Gao,
        and~Xinbo~Gao,~\IEEEmembership{Fellow,~IEEE}}
\markboth{IEEE TRANSACTIONS ON MULTIMEDIA}
{Shell \MakeLowercase{\textit{et al.}}: Bare Demo of IEEEtran.cls for IEEE Journals}
% The only time the second header will appear is for the odd numbered pages
% after the title page when using the twoside option.
%
% *** Note that you probably will NOT want to include the author's ***
% *** name in the headers of peer review papers.                   ***
% You can use \ifCLASSOPTIONpeerreview for conditional compilation here if
% you desire.

% If you want to put a publisher's ID mark on the page you can do it like
% this:
%\IEEEpubid{0000--0000/00\$00.00~\copyright~2015 IEEE}
% Remember, if you use this you must call \IEEEpubidadjcol in the second
% column for its text to clear the IEEEpubid mark.

% use for special paper notices
%\IEEEspecialpapernotice{(Invited Paper)}

% make the title area
\maketitle

% As a general rule, do not put math, special symbols or citations
% in the abstract or keywords.
\begin{abstract}
Weakly supervised LiDAR semantic segmentation has made significant strides with limited labeled data. However, most existing methods focus on the network training under weak supervision, while efficient annotation strategies remain largely unexplored. To tackle this gap, we implement LiDAR semantic segmentation using scatter image annotation, effectively integrating an efficient annotation strategy with network training. Specifically, we propose employing scatter images to annotate LiDAR point clouds, combining a pre-trained optical flow estimation network with a foundation image segmentation model to rapidly propagate manual annotations into dense labels for both images and point clouds. Moreover, we propose ScatterNet, a network that includes three pivotal strategies to reduce the performance gap caused by such annotations. Firstly, it utilizes dense semantic labels as supervision for the image branch, alleviating the modality imbalance between point clouds and images. Secondly, an intermediate fusion branch is proposed to obtain multimodal texture and structural features. Lastly, a perception consistency loss is introduced to determine which information needs to be fused and which needs to be discarded during the fusion process. Extensive experiments on the nuScenes and SemanticKITTI datasets have demonstrated that our method requires less than 0.02\% of the labeled points to achieve over 95\% of the performance of fully-supervised methods. Notably, our labeled points are only 5\% of those used in the most advanced weakly supervised methods.
\end{abstract}

% Note that keywords are not normally used for peerreview papers.

% For peer review papers, you can put extra information on the cover
% page as needed:
% \ifCLASSOPTIONpeerreview
% \begin{center} \bfseries EDICS Category: 3-BBND \end{center}
% \fi
%
% For peerreview papers, this IEEEtran command inserts a page break and
% creates the second title. It will be ignored for other modes.
\IEEEpeerreviewmaketitle

\section{Introduction}
% The very first letter is a 2 line initial drop letter followed
% by the rest of the first word in caps.
%
% form to use if the first word consists of a single letter:
% \IEEEPARstart{A}{demo} file is ....
%
% form to use if you need the single drop letter followed by
% normal text (unknown if ever used by the IEEE):
% \IEEEPARstart{A}{}demo file is ....
%
% Some journals put the first two words in caps:
% \IEEEPARstart{T}{his demo} file is ....
%
% Here we have the typical use of a "T" for an initial drop letter
% and "HIS" in caps to complete the first word.

\IEEEPARstart{I}{n} recent years, weakly supervised LiDAR semantic segmentation has attracted considerable research attention, achieving performance comparable to fully supervised methods while using fewer annotated points \cite{10487013}. However, current methods primarily concentrate on network design and overlook the integration of network training with cost-effective annotation strategies. In this paper, we address this oversight by combining the two: first introducing a new annotation strategy, and then developing a multimodal training approach that directly leverages these annotations.

\begin{figure}[!t]
\centering
\includegraphics[width=0.85\linewidth]{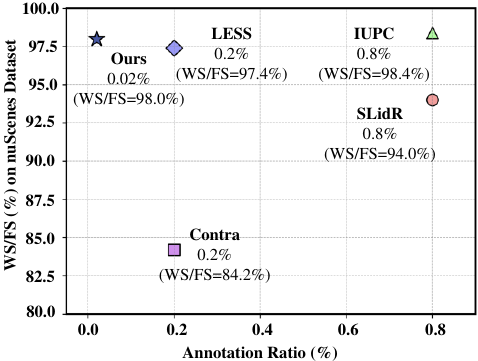}
\caption{Comparison of our method with  SLidR \cite{sautier2022image}, IUPC \cite{sun2024image}, LESS \cite{liu2022less}, and Contra \cite{hou2021exploring} on the nuScenes dataset. WS/FS denotes their relative performance compared to the fully supervised method.}
\label{fig:1}
\end{figure}

\textbf{Regarding annotation strategy}, current methods can be categorized based on different levels of supervision, including partially labeled points \cite{liu2021one,yang2022mil,zhang2021weakly}, sub-cloud level annotations \cite{wei2020multi}, and scene-level annotations \cite{wei2020multi,li2024multi}. Among several weak labeling schemes, the partially labeled point clouds scheme provides the best balance between annotation cost and segmentation performance \cite{hu2022sqn,liu2022less,liu2023cpcm}, which is also the focus of our paper. In partially annotated point clouds, although existing methods require only a tiny part of points to be labeled (e.g 0.1\%), the annotation process still incurs considerable costs. \textbf{Firstly}, the amount of manual annotation is still substantial. For example, given a frame from LiDAR containing 100,000 points, even a labeling ratio of 0.1\% requires hundreds of click-based annotations. \textbf{Secondly}, the annotation process is cumbersome. Switching viewpoints and zooming in and out of the entire scene is cumbersome and does not significantly save time. To address these challenges, we introduce a strategy called \textbf{Scatter Image Annotation}. Specifically, we utilize GMFlowNet [12], a pre-trained optical flow estimation model, and SAM [13], a foundation image segmentation model, to minimize the annotation workload. This approach propagates manual annotations from sparse to dense, drastically reducing the annotation cost compared to existing methods, as depicted in Fig. \ref{fig:1}. For instance, our method effectively reduces manual annotation costs by 90\% compared to LESS \cite{liu2022less}. Additionally, to simplify the annotation process, we transition from point-based to pixel-based annotation. This change allows annotators to annotate each class with just 1-5 mouse clicks on a 2D plane, eliminating the complexities associated with 3D rotation. It is important to note that while \cite{wang2020weakly} uses pixel-based annotation, it employs dense annotation instead of our more efficient sparse pixel approach, making it more time-consuming.

\begin{figure}[ht]
\centering
\includegraphics[width=0.8\linewidth]{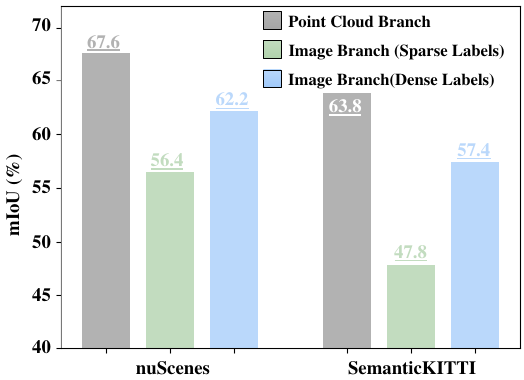}
\caption{Mitigating modality imbalances using dense semantic labels. We train the point cloud branch and two variants of the image branch, one with sparse labels and the other with dense ones, separately in a fully supervised manner on the SemanticKITTI \cite{behley2019semantickitti} and nuScenes \cite{behley2019semantickitti} datasets. The mIoU is the evaluation metric for semantic segmentation; higher values indicate greater accuracy.}
\label{fig:2}
\end{figure}

\begin{figure}[ht]
  \centering
  \includegraphics[width=0.95\linewidth]{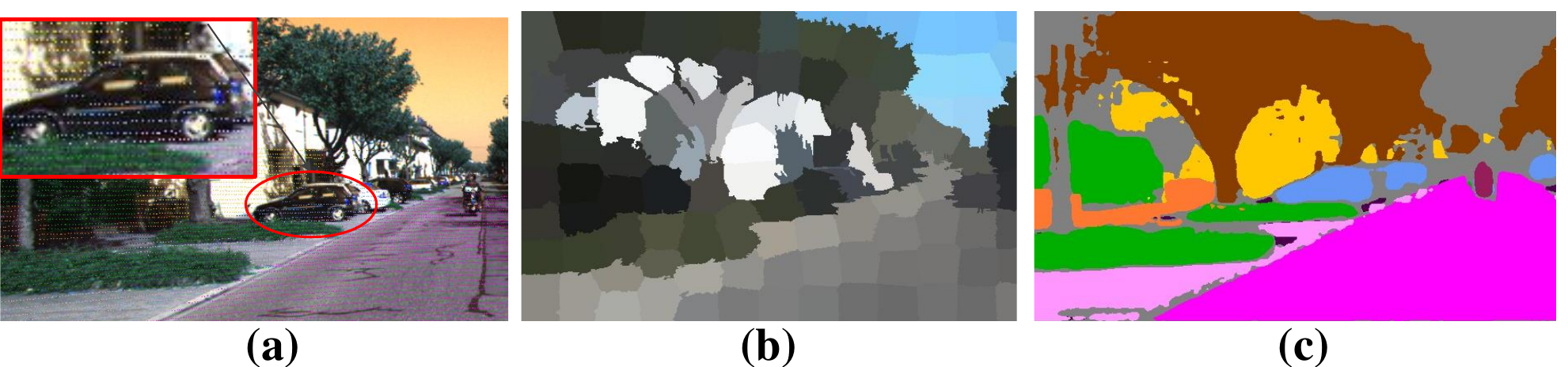} % 图片的文件名和路径
  \caption{Comparison of supervision labels for the image branch. (a) Sparse semantic labels: Projecting point cloud labels onto images. (b) Superpixel labels: Using the SLIC \cite{achanta2012slic} to group pixels into superpixels. (c) Our method: Using SAM \cite{kirillov2023segment} to generate dense semantic labels from manual annotations.} % 图片的标题
  \label{fig:3} % 图片的标签，用于引用
\end{figure}

% \begin{figure}[ht]
%   \centering
%   \includegraphics[width=0.98\linewidth]{figures/figure_4.pdf} % 图片的文件名和路径
%   \caption{Comparison of ScatterNet with most multimodal methods. (a) Most multimodal fusion methods: Fusion occurs only at the feature encoding or decoding stages. (b) Our method: Uses an independent fusion branch that integrates at both the feature encoding and decoding stages. $\operatorname{O}$ indicates classification output.} % 图片的标题
%   \label{fig:4} % 图片的标签，用于引用
% \end{figure}

\par \textbf{Regarding training}, we train the network using dense multimodal labels from scatter image annotations, and leverage the prediction consistency between multimodal features as an additional constraint for network training. However, due to imbalanced capabilities, the performance of the image branch is significantly weaker than that of the point cloud branch (see Fig. \ref{fig:2}, where the point cloud branch outperforms the image branch across all datasets). Therefore, naively enforcing consistency between the outputs of the two modalities will degrade performance. To address this issue, we propose \textbf{ScatterNet}, which includes three key strategies. \textbf{Firstly}, we employ dense semantic labels as the supervision for the image branch. Previous image branches typically use either sparse labels, as depicted in Fig. \ref{fig:3}(a), or superpixels, as depicted in Fig. \ref{fig:3}(b), for supervision. The former is deficient in dense label information, whereas the latter is devoid of semantic information. To generate dense semantic labels, we employ SAM \cite{kirillov2023segment} to propagate sparse image annotations into dense semantic labels, as shown in Fig. \ref{fig:3}(c). Subsequent analysis and the experiment indicate that using these dense semantic labels effectively alleviates modality imbalances. For example, on the SemanticKITTI dataset, the mIoU gap has narrowed from 16.0\% to 6.4\%, as shown in Fig. \ref{fig:2}. \textbf{Secondly}, we propose an intermediate fusion branch to integrate point cloud and image features. Most existing multimodal fusion networks either project point clouds onto the image plane during encoding \cite{zhuang2021perception,wu2023cross} or lift image features into the point cloud space during decoding \cite{yan20222dpass}. Our approach leverages the advantages of both methods by employing an independent fusion branch, enabling feature fusion throughout both the encoding and decoding stages. This allows us to maximize the exploitation of texture and structural information from multimodal data. \textbf{Thirdly}, we propose a perception consistency loss to determine which information needs to be fused and which needs to be discarded during the fusion process.

\par In summary, combining scatter image annotation with multimodal training efficiently covers more underrepresented instances within a limited labeling budget. The contributions of our work can be summarized as follows:

\begin{itemize}
  \item [1)]
  We propose a weakly supervised semantic segmentation approach using scatter image annotation. This method synergizes an efficient annotation strategy with network training, achieving performance comparable to that of fully supervised methods. On the nuScenes dataset, with only 0.02\% of the labeled data, our method achieves 98.0\% of the performance of fully supervised methods (see Fig. \ref{fig:1}).
  \item [2)]
  We propose a pixel-based weakly supervised annotation strategy named Scatter Image Annotation. This strategy employs a pre-trained optical flow estimation network and a foundation image segmentation model to propagate manual annotations into dense labels for both images and point clouds, thereby simplifying and accelerating the traditional point-based annotation.
  \item [3)]
  We propose ScatterNet, which utilizes dense semantic labels to supervise the image branch and introduces an intermediate fusion branch to acquire multimodal texture and structural features, effectively mitigating the modality imbalance. Additionally, the network incorporates a perception consistency loss to determine which information to fuse and which to discard during the fusion process.
 
\end{itemize}

\section{Related Work}
\subsection{Weakly Supervised Annotation Strategy}
\begin{figure}[!t]
  \centering
  \includegraphics[width=0.95\linewidth]{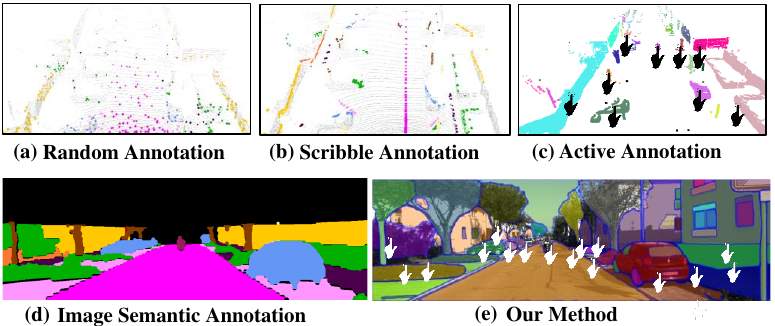} % 图片的文件名和路径
  \caption{Weakly supervised annotation strategy:
\textbf{(a) Random annotation}: Randomly selects a small subset of points for annotation. This approach has difficulty covering infrequent small objects, such as distant fences and cyclists.
\textbf{(b) Scribble annotation}: Line-based scribbles, requiring just two clicks (start and end points), allow for quick marking of large geometric areas. However, this method is less effective for smaller objects such as pedestrians.
\textbf{(c) Active annotation}: Ground detection is used to separate ground and non-ground points, after which the non-ground points are clustered into multiple point cloud subsets. Each subset is then labeled by annotators.
\textbf{(d) Image semantic annotation}: Image semantic segmentation labels are mapped to 3D space for point cloud annotation. This method is more convenient compared to point-based annotation, as it does not require rotating the 3D space. However, dense pixel-based annotation is still time-consuming. \textbf{(e) Our method (Scatter Image Annotation)}: Just 1-5 mouse clicks on an image are required to annotate each instance. This sparse, pixel-based approach provides ease of interaction and reduces annotation costs.} % 图片的标题
  \label{fig:5} % 图片的标签，用于引用
\end{figure}

Previous weakly supervised annotation strategies have primarily focused on indoor datasets, such as ScanNet-v2 \cite{dai2017scannet} and S3DIS \cite{armeni2017joint}, including studies \cite{yang2022mil,liu2021one,liu2023cpcm,hu2022sqn,wu2022dual,li2024multi,pan2024less,10487013}. In these datasets, point clouds are sampled from high-quality reconstructed meshes, resulting in dense and evenly distributed data. However, in outdoor settings, the highly imbalanced frequency and size of different categories have made \textbf{random annotation} \cite{liu2021one,hu2022sqn,zhang2021perturbed} challenging, resulting in infrequent small objects being rarely labeled. For example, distant fences and cyclists could not be properly annotated, as shown in Fig. \ref{fig:5}(a). To address these challenges, Ozan Unal et al. \cite{unal2022scribble} propose \textbf{scribble annotation}. This approach uses line-based scribbles to annotate point clouds, allowing faster labeling of large-scale geometric categories such as roads, sidewalks, buildings, and fences. As shown in Fig. \ref{fig:5}(b), distant fences and roads can be effectively annotated. However, it is less effective for
smaller objects like pedestrians. Recent studies \cite{sun2024image,liu2022less} have explored \textbf{active annotation}. This method first uses a ground detection algorithm to divide the point cloud into ground and non-ground sections, then clusters the non-ground points into multiple point cloud subsets. Each subset is then annotated, as depicted in Fig. \ref{fig:5}(c). However, these point-based methods require frequent rotation of the 3D view to label rare or small objects. To overcome this challenge, a pixel-based approach called \textbf{image semantic annotation} is introduced. These methods \cite{wang2020weakly,genova2021learning} use dense image semantic segmentation labels for point cloud segmentation, as shown in Fig. \ref{fig:5}(d). However, dense pixel-based annotation remains time-consuming. To reduce the cost of dense pixel-based annotation, we propose \textbf{scatter image annotation}. With this method, annotating each instance requires only 1-5 mouse clicks on a 2D image, as shown in Fig. \ref{fig:5}(e), making the process simple and highly efficient.

\subsection{Multimodal Point Cloud Segmentation}
\par Multimodal methods aim to fuse information from two complementary sensors, utilizing the benefits of both cameras and LiDAR. However, because point clouds and images exist in different metric spaces, it is difficult to fuse them directly using the convolution or attention mechanism. To solve this problem, one solution is to project point clouds in a 2D space during encoding \cite{liang2019multi,liang2018deep,zhuang2021perception,wu2023cross}, thus aligning the point cloud and the image in the same coordinate system. This allows the direct application of existing image segmentation methods to the projected point cloud. Moreover, the shared coordinate system between the projected point cloud and the image facilitates the effective fusion of intermediate feature layers. However, projecting point clouds onto a 2D space results in the neglect of their 3D structural information. To preserve this structural information, another approach \cite{jaritz2020xmuda,yan20222dpass} involves lifting the image features to a 3D space. However, since not all image pixels correspond to the point cloud, these methods compromise the texture features of the image. Our approach integrates the strengths of both projection and lifting methods by employing independent fusion branches. This enables feature fusion during both encoding and decoding stages, maximizing the utilization of both texture and structural information from multimodal data.

\subsection{Weakly-supervised Point Cloud Segmentation}

\par Point cloud segmentation \cite{10330760,9913730} have made great strides with fully annotated datasets, but creating them requires substantial effort in point-level labeling. To deal with the annotation bottleneck, some recent work has delved deeply into the field of weakly supervised semantic segmentation \cite{wang2023survey}. These methods can be broadly categorized into two types: \textit{pseudo-labeling methods} and \textit{consistency regularization methods}. \textbf{Pseudo-labeling} methods explore by predicting pseudo-labels for unlabeled points. MPRM \cite{wei2020multi} trains segmentation models on sub-cloud labels and uses class activation mapping \cite{zhou2016learning} to generate pseudo-labels for entire sub-clouds, which are then used to train the final model. OTOC \cite{liu2021one} enhances the quality of pseudo-labels via iterative self-training processes. SQN \cite{hu2022sqn} optimizes the use of sparse labels by incorporating geometric priors. However, utilizing pseudo-labels for model iteration optimization often suffers from the influence of unreliable pseudo-labels, resulting in poor segmentation performance. Therefore, \textbf{consistency regularization} reduces the impact of unreliable labels by imposing consistency constraints on the predicted distributions. For instance, HybridCR \cite{li2022hybridcr} utilizes hybrid contrast regularization to encourage anchor points to match positive samples  while being dissimilar to negative points. Wu \textit{et al.} \cite{wu2022dual} utilize a sufficient smoothness constraint by applying a large amount of unlabeled points for weakly supervised point cloud segmentation. Li \textit{et al.} \cite{li2024multi} and \textit{Sun et al.} \cite{sun2024image} fully exploit the point feature affinities of multiple data modalities to limit multimodal features. Instead of focusing primarily on network design, as previous methods do, we propose a weakly supervised semantic segmentation approach that combines an efficient annotation strategy with network training. This method not only simplifies the annotation process but also significantly enhances the efficiency and effectiveness of network training.

\section{Proposed Method}

\begin{figure*}[ht]
  \centering
  \includegraphics[width=0.90\linewidth]{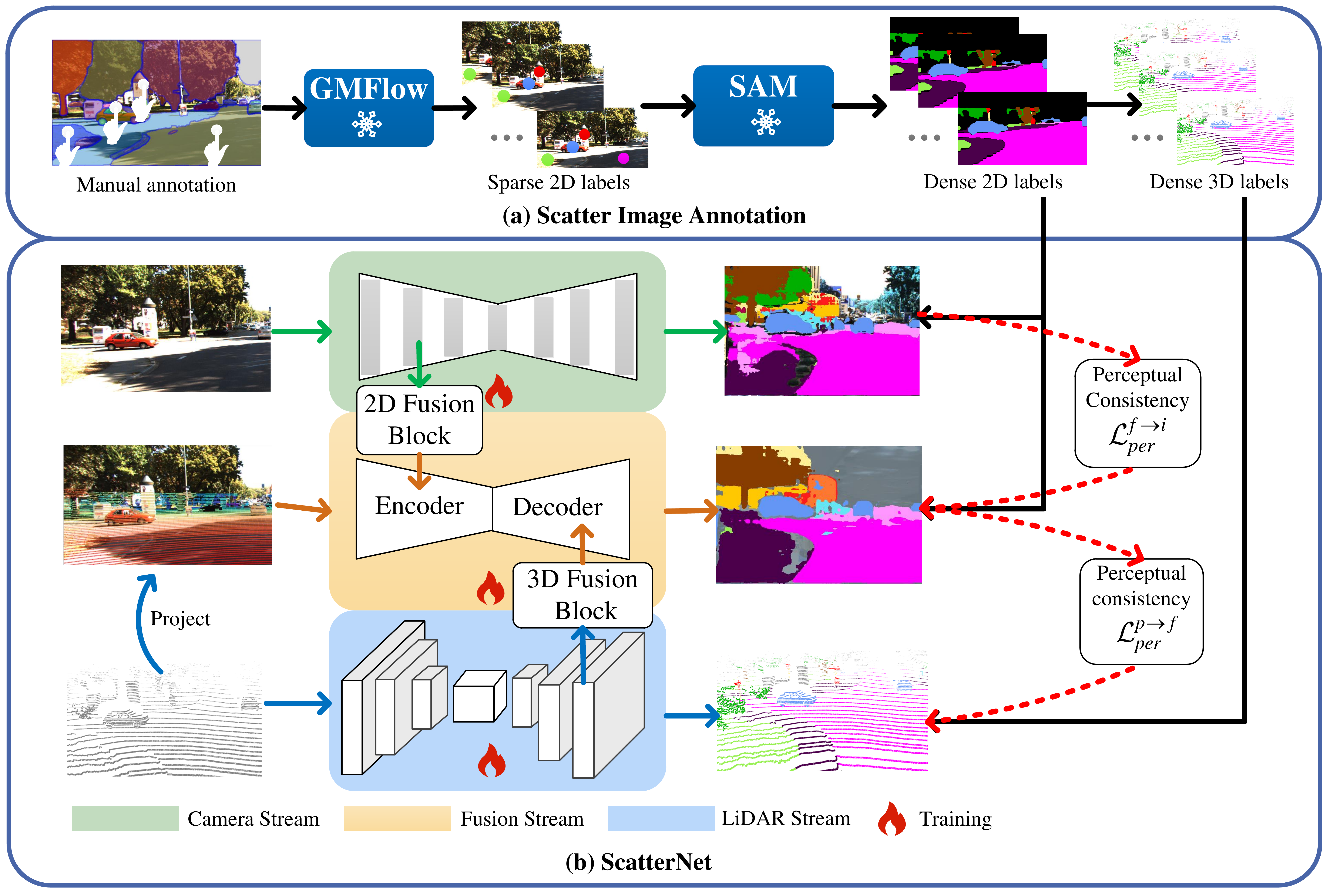} 
  \caption{\textbf{Overview of our method.} (a) Scatter Image Annotation: Using GMFlowNet \cite{xu2022gmflow} and SAM \cite{kirillov2023segment}, manual annotations are propagated to dense labels. (b) ScatterNet: The network consists of three branches: \textit{Camera Stream}, \textit{Fusion Stream}, and \textit{LiDAR Stream}. The \textit{Camera Stream} is used for the extraction of image features; the \textit{LiDAR Stream} handles the processing of point cloud features; while the \textit{Fusion Stream} is used to merge the features extracted by the \textit{Camera} and \textit{LiDAR streams}. The loss of perceptual consistency consists of two parts: $\mathcal{L}_{per}^{f\to i}$ represents the consistency between \textit{Camera} and \textit{Fusion} Streams, and $\mathcal{L}_{per}^{p\to f}$ represents the consistency between \textit{LiDAR} and \textit{Fusion streams}.} % 图片的标题
  \label{fig:6} % 图片的标签，用于引用
\end{figure*}

\subsection{Overview of Our Method}
\par Our goal is to achieve LiDAR semantic segmentation comparable to fully supervised methods using only a tiny part of labels. To this end, we have synergized an efficient annotation strategy with network training to develop a simple yet effective framework. This framework encompasses scatter image annotation strategy and the training network, ScatterNet, as shown in Fig. \ref{fig:6}. The specific process is as follows: 

\par For the annotation strategy, we propose the scatter image annotation to generate dense 2D and 3D labels from manual annotations, as illustrated in Fig. \ref{fig:6} (a). First, sparse 2D labels are acquired through interframe label propagation using the pre-trained GMFlowNet \cite{xu2022gmflow}. Subsequently, these labels are densified through intraframe label propagation using SAM \cite{kirillov2023segment}. Ultimately, the dense 2D labels are mapped into 3D space to yield dense 3D labels. For further details, see Sec. \ref{sec: 3.2}. 

\par For network training, we use dense labels extracted from Scatter Image annotation for supervision, as depicted in Fig. \ref{fig:6} e(b). Specifically, we develop a model named ScatterNet, which includes an independent \textit{Fusion Stream} designed to integrate features from the \textit{Camera} and \textit{LiDAR Streams}. In this process, the \textit{Camera} and \textit{Fusion Streams} are supervised using dense 2D labels, while the \textit{LiDAR Stream} utilizes dense 3D labels. Additionally, to determine which information needs to be fused, we introduce a perception consistency loss. It is important to note that during testing, we exclusively use the \textit{LiDAR Stream}, thus facilitating deployment without requiring image input. For additional details, see Sec. \ref{sec: 3.3}.

%ScatterNet comprises three branches: \textit{Camera}, \textit{Fusion} and \textit{LiDAR Streams}, as depicted in Fig. \ref{fig:6}(b). Both the \textit{Camera Stream} and \textit{Fusion Stream} are supervised using dense 2D labels, while the \textit{LiDAR Stream} utilizes dense 3D labels. Furthermore, a perception consistency loss is introduced to determine which information needs to be fused. It is important to note that during testing, we use only the \textit{LiDAR Stream}, thus facilitating deployment without requiring image input. For additional details, see Sec. \ref{sec: 3.3}.

\subsection{Scatter Image Annotation}
\label{sec: 3.2}

\begin{figure*}[ht]
\centering
\setlength{\belowcaptionskip}{-3mm}
\includegraphics[width=0.90\linewidth]{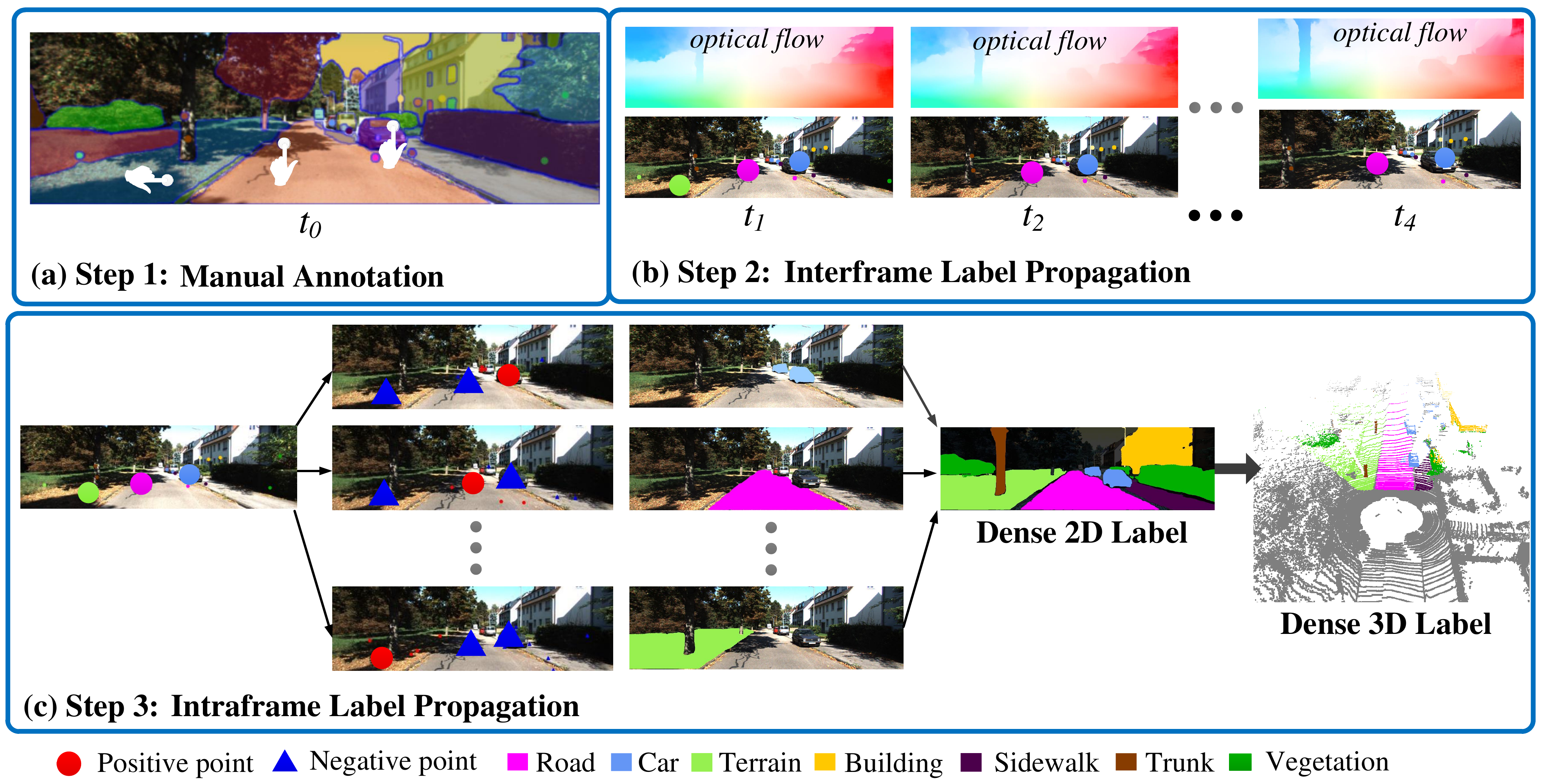}
\vspace{-0.0cm}
\caption{Steps for Scatter Image Annotation: \textbf{(a) Step 1}: Manual Annotation. \textbf{(b) Step 2}: Interframe Label Propagation. The optical flow estimation network is utilized to propagate the labels from the first step to the next four successive frames. \textbf{(c) Step 3}: Intraframe Label Propagation. The image segmentation model is used to propagate the sparse labels generated in the second step into dense labels.}
\label{fig:7}
\end{figure*}

\par In typical point cloud scenarios, such as autonomous driving, cameras are often used in conjunction with auxiliary LiDAR. By synchronizing and calibrating the LiDAR with the camera, we establish a clear mapping between point cloud and image data. This inherent physical spatial relationship allows us to label point clouds through sparse manual annotations on images. However, the sparse labels are insufficient to provide adequate supervision for training. To address this issue, we introduce a new strategy called Scatter Image Annotation, which uses pre-trained optical flow and image segmentation models to propagate sparse labels into dense ones. Specifically, First, we detail the implementation steps of this strategy. We then apply and evaluate its effectiveness on two widely used datasets, SemanticKITTI \cite{behley2019semantickitti} and nuScenes \cite{caesar2020nuscenes}, focusing on the quantity and quality of the annotations. Finally, we explore how to apply this strategy in other data collection scenarios.

%To address this issue, we first introduce a strategy known as Scatter Image Annotation. We then apply and assess this strategy on two widely used datasets, SemanticKITTI \cite{behley2019semantickitti} and nuScenes \cite{caesar2020nuscenes},  focusing on both the quantity and quality of the annotations. Finally, we explore how to apply this strategy in other data collection scenarios.

% In typical point cloud scenarios, such as autonomous driving, cameras are often used in conjunction with auxiliary LiDAR. By synchronizing and calibrating the LiDAR and camera, we establish a clear mapping between 3D and 2D data through point-to-pixel transformation. By harnessing this inherent physical spatial relationship, we introduce an annotation strategy termed Scatter Image Annotation, which is applied to two widely used datasets, SemanticKITTI \cite{behley2019semantickitti} and nuScenes \cite{caesar2020nuscenes}.

\subsubsection{Steps for Scatter Image Annotation}
\par The annotation strategy is divided into three steps, as illustrated in Fig. \ref{fig:7}.

\textbf{Step 1: Manual Annotation.} Annotators apply 1 to 5 scatter annotations to each instance based on the pixel area size in the image (e.g., roads typically require 5 pixel-level annotations, while pedestrians require 1). To help annotators quickly identify distinct instances, we utilize SAM \cite{kirillov2023segment} for the preliminary segmentation of unlabeled images, as shown in Fig. \ref{fig:7}(a). In this paper, we simulate the manual annotation process using the semantic segmentation labels provided by the dataset. Specifically, we project the point cloud labels onto images. Subsequently, we randomly retain 1-5 representative labels for each class.

\par \textbf{Step 2: Interframe Label Propagation.} To reduce the cost of manual annotation, we use interframe propagation to automatically label consecutive frames. Given the temporal continuity of images in autonomous driving scenarios, we utilize a pre-trained optical flow estimation model, GMFlowNet \cite{xu2022gmflow}, to predict the positional shifts of manual annotations. This allows us to extend the manual annotations from the initial frame to the subsequent four frames. The specific reasons for selecting four frames will be detailed in Sec. \ref{Sec:3.2.4}. It is important to note that the optical flow estimation model was trained on the synthetic dataset Sintel \cite{Butler:ECCV:2012}, ensuring that this approach does not incur additional manual labeling costs.

%Leveraging the temporal continuity of images in autonomous driving scenarios, we employ a pre-trained optical flow estimation model, GMFlowNet \cite{xu2022gmflow}, to propagate the manual annotations from the initial frame to the next four successive frames, as depicted in Fig. \ref{fig:7}(b). Based on experiments, we find that as long as the number of propagated frames is kept a certain range, this method not only maintains labeling accuracy but also significantly reduces the labeling cost. Importantly, the optical flow estimation model has been trained on a synthetic dataset, Sintel \cite{Butler:ECCV:2012}, ensuring that this method does not incur additional manual labeling costs.

\par \textbf{Step 3: Intraframe Label Propagation.} Due to the sparse labels obtained through manual annotation and interframe propagation being insufficient to provide adequate supervision, we employ intraframe propagation to densify these sparse labels. Specifically, we initially use a foundation image segmentation model, SAM \cite{kirillov2023segment}, to generate preliminary segmentation masks for each category. For example, to generate a mask for cars, we consider point labels of cars as positive points, while point labels of other classes (such as trees, grass, buildings, etc.) are considered negative points. Then, we feed the positive and negative point labels along with the image into the SAM to produce the segmentation mask for car. Subsequently, we generate segmentation masks for each class and merge all the segmentation masks to obtain the final dense 2D label. Finally, we map the dense 2D labels to 3D space, thus producing dense 3D labels, as illustrated in Fig. \ref{fig:7}(c).

%Using SAM \cite{kirillov2023segment}, a foundation image segmentation model, we initially generate segmentation masks for each class. For example, to generate a mask for cars, we consider point labels of cars as positive points, while point labels of other classes (such as trees, grass, buildings, etc.) are considered negative points. Then, we feed the positive and negative point labels along with the image into the SAM \cite{kirillov2023segment} to produce the segmentation mask for car. Subsequently, we generate segmentation masks for each class and merge all the segmentation masks to obtain the final dense 2D label. Finally, we utilize the inherent physical spatial relationship between the camera and LiDAR to map the dense 2D labels to 3D space, thus producing dense 3D labels, as illustrated in Fig. \ref{fig:7}(c).

\subsubsection{Count of Scatter Image Annotation}

\begin{figure}[htb]
  \centering
  \includegraphics[width=0.90\linewidth]{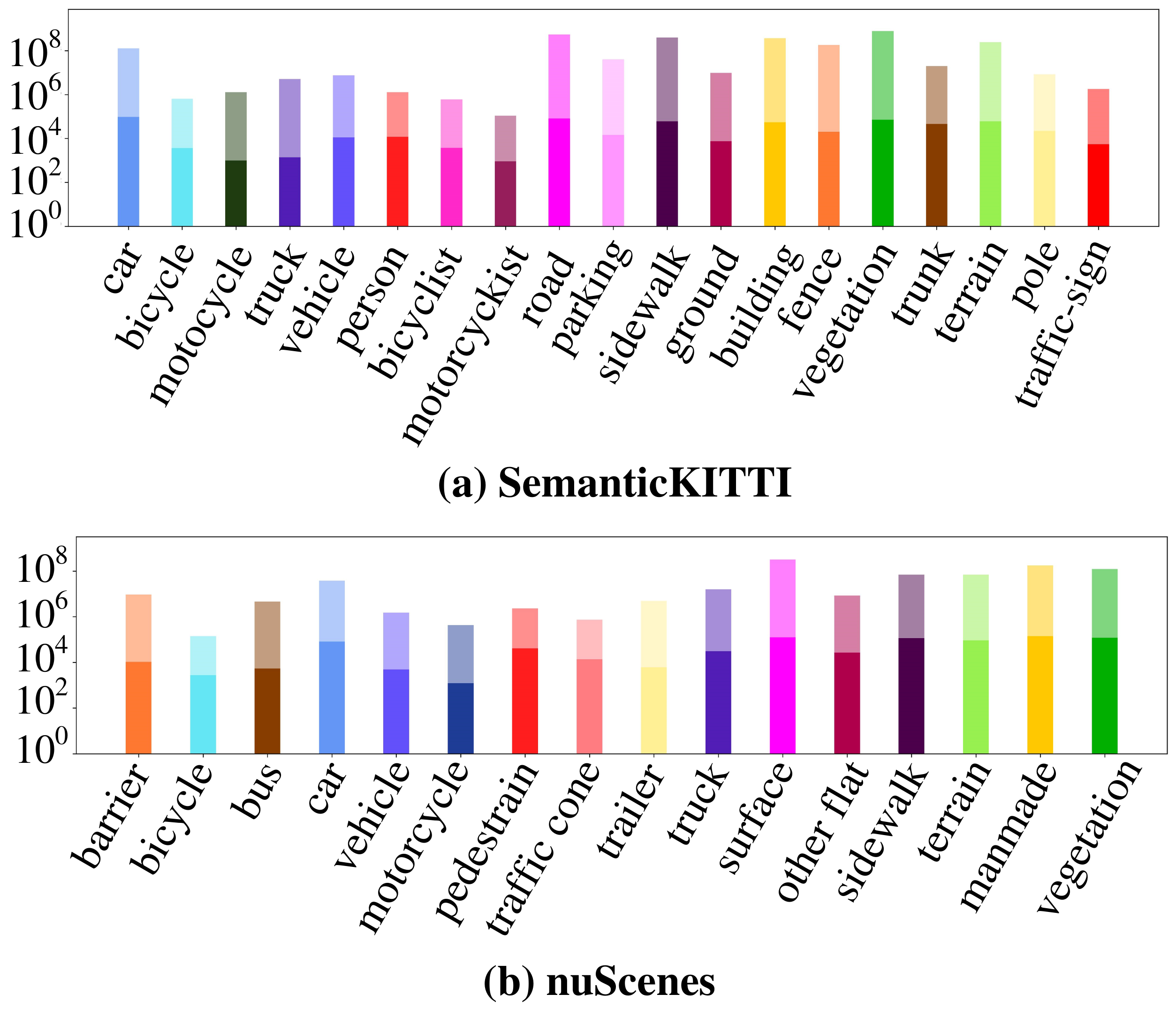} % 图片的文件名和路径
  \caption{Counting Scatter Image Annotation. Light colors represent the number of labels in SemanticKITTI and nuScenes, while dark colors represent the number of labels using \textbf{Scatter Image Annotation}. (a) The number of labeled points on SemanticKITTI \cite{behley2019semantickitti} in log scale; (b) The number of labeled points on nuScenes \cite{caesar2020nuscenes} in log scale.} % 图片的标题
  \label{fig:8} % 图片的标签，用于引用
\end{figure}

\par We apply the scatter image annotation strategy to both the SemanticKITTI and nuScenes datasets and document the number of manual annotations. The SemanticKITTI dataset is equipped with a 64-beam LiDAR and contains 2.848 billion points. In this dataset, we have annotated 0.11 million points, which is only 0.004\% of the total. To facilitate a smooth transition to weakly supervised LiDAR semantic segmentation, we have retained the same 19 categories. The distribution of the category labels is shown in Fig. \ref{fig:8}(a).  For the nuScenes dataset, which uses a 32-beam LiDAR and contains 1.185 billion points, we have annotated 0.23 million points, comprising just 0.02\% of the total. To encourage a smooth transition to weakly supervised LiDAR semantic segmentation research, we have retained the same 17 categories. The distribution of category labels can be seen in Fig. \ref{fig:8}(b). The statistical results from both datasets demonstrate that our annotation strategy significantly reduces the need for manual annotations.

\subsubsection{Evaluation of Scatter Image Annotation}

\par
\begin{figure}[htbp]
  \centering
  \includegraphics[width=0.90\linewidth]{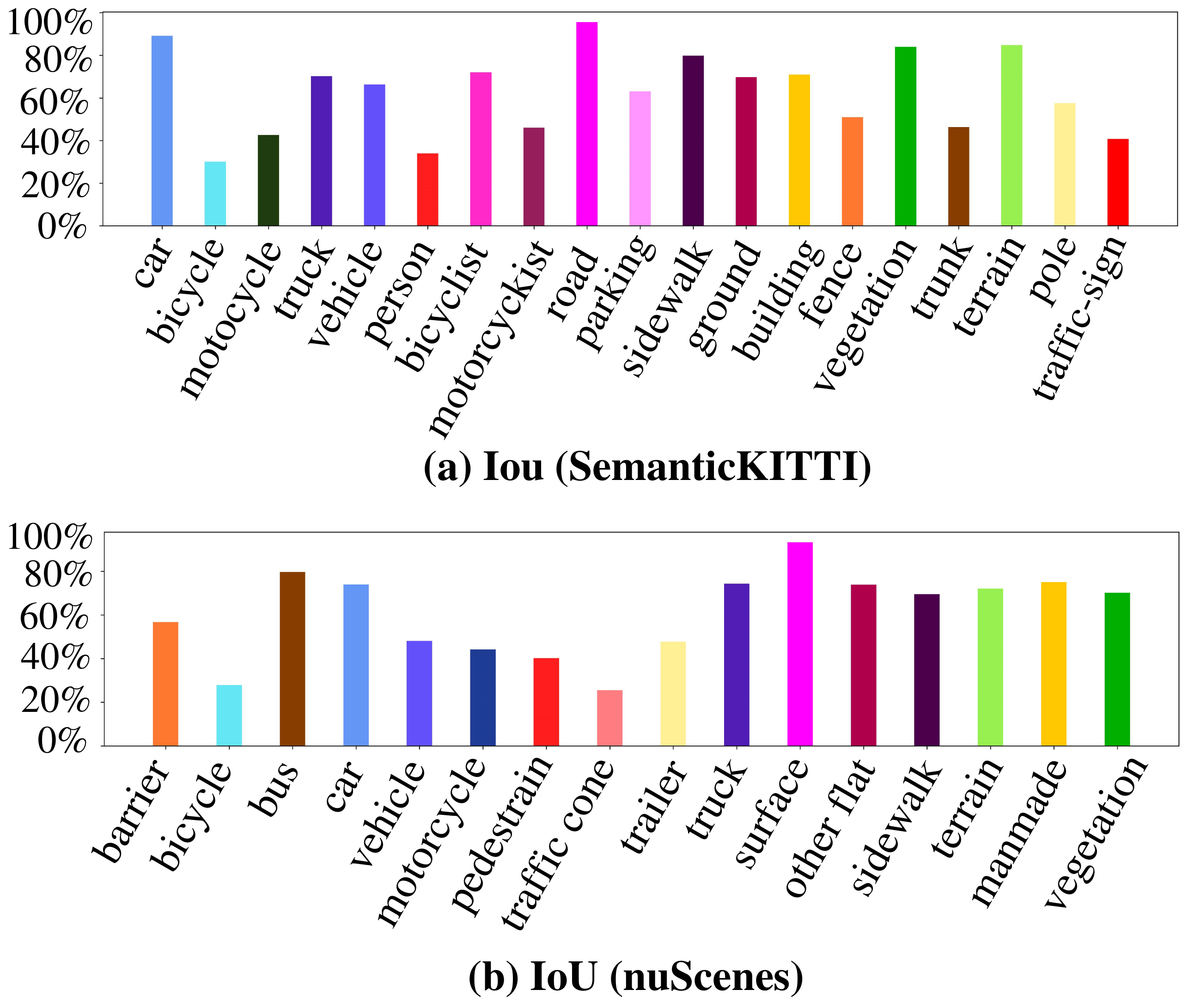} % 图片的文件名和路径
  \caption{Dense label IoU for each class. (a) Results on SemanticKITTI \cite{behley2019semantickitti}; (b) Results on nuScenes \cite{caesar2020nuscenes}.} % 图片的标题
  \label{fig:10} % 图片的标签，用于引用
\end{figure}

\begin{figure}[htbp]
  \centering
  \includegraphics[width=0.95\linewidth]{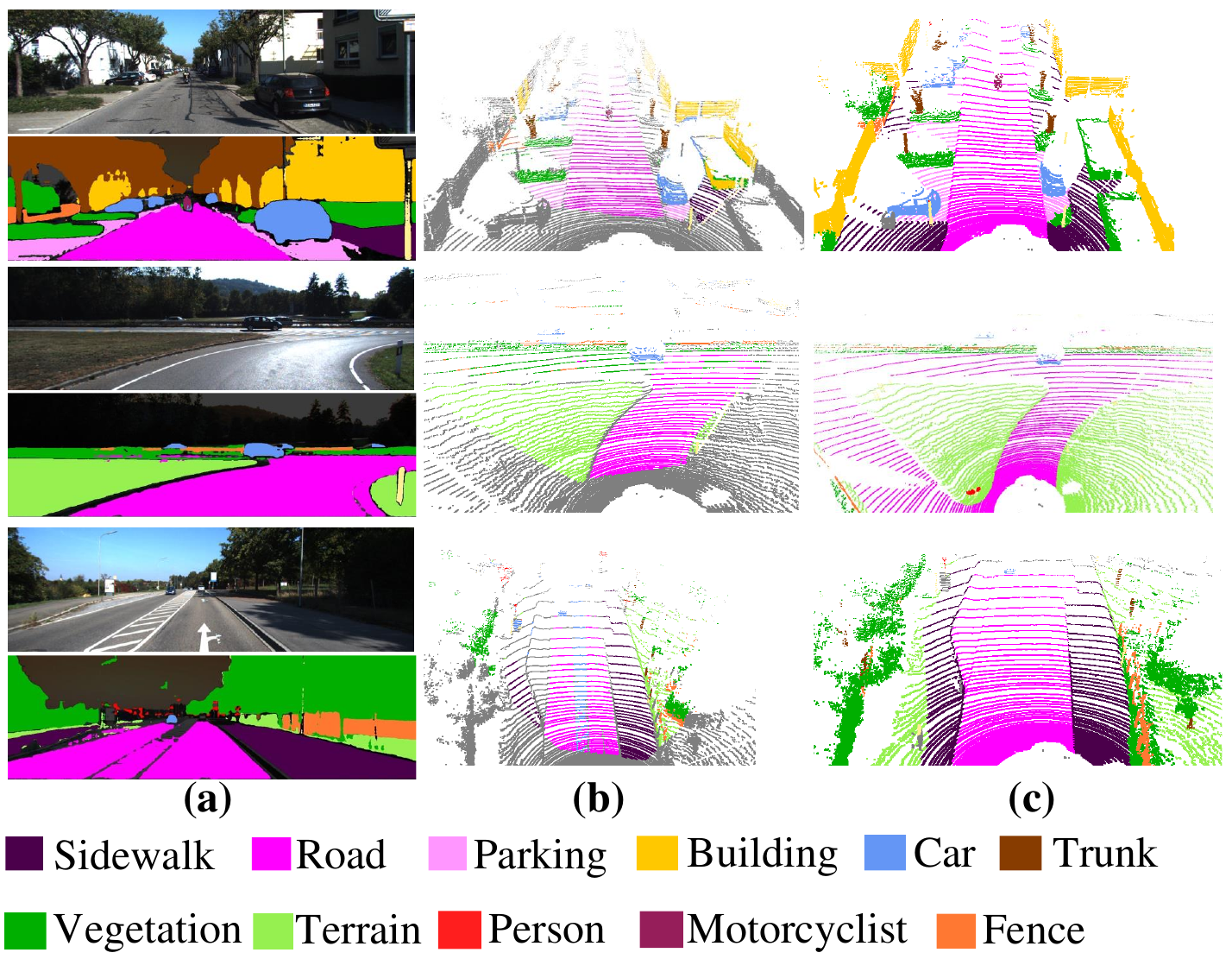} % 图片的文件名和路径
  \caption{Visualizing the results of Step 3 in scatter image annotation on SemanticKITTI \cite{behley2019semantickitti}. (a) Dense 2D labels, (b) Dense 3D labels. (c) Ground truth labels.} % 图片的标题
  \label{fig:11} % 图片的标签，用于引用
\end{figure}

\begin{figure}[htbp]
  \centering
  \includegraphics[width=0.95\linewidth]{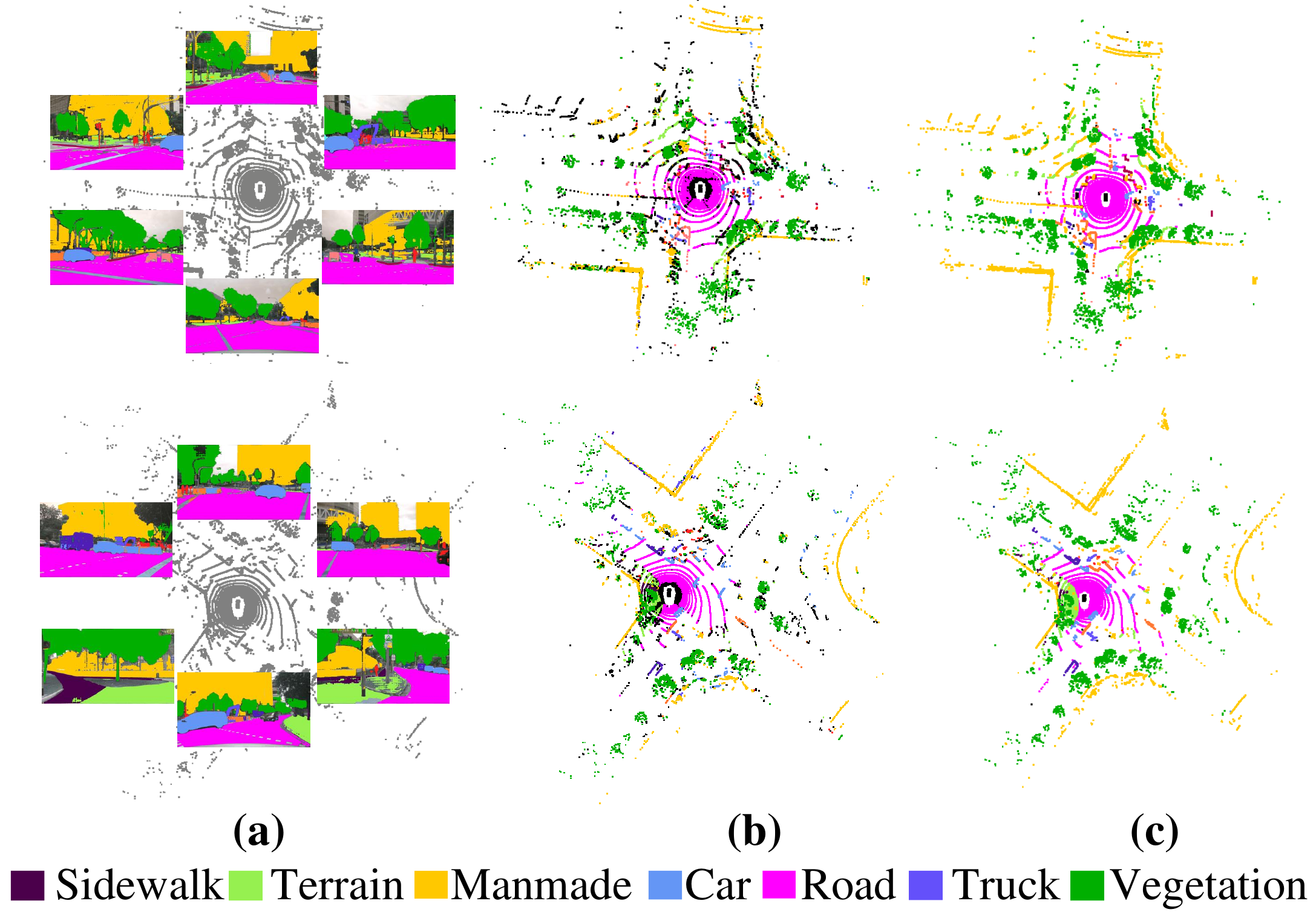} % 图片的文件名和路径
  \caption{Visualizing the results of Step 3 in scatter image annotation on nuScenes \cite{caesar2020nuscenes}. (a) Dense 2D labels, (b) Dense 3D labels. (c) Ground truth labels.} % 图片的标题
  \label{fig:12} % 图片的标签，用于引用
\end{figure}

 To evaluate the quality of Scatter Image Annotation, we conduct both quantitative and qualitative evaluations of the dense labels. For quantitative analysis, we used Intersection over Union (IoU) and Mean Intersection over Union (mIoU) as evaluation metrics. Since our annotation strategy is pixel-based, we only assess the dense labels within the camera's field of view. On the SemanticKITTI and nuScenes datasets, we calculate the mIoU for dense labels, obtaining results of 68\% and 77\% respectively. Additionally, we calculate the IoU for different categories, with detailed information shown in Fig. \ref{fig:10}. In addition to quantitative analysis, we also conduct qualitative evaluations. For instance, Fig. \ref{fig:11} displays the dense 2D and 3D labels from the SemanticKITTI dataset. Since this dataset is only equipped with a front-view camera, it only shows the point clouds of the foreground. In contrast, as shown in Fig. \ref{fig:12}, the nuScenes dataset includes six cameras, enabling the visualization of the entire panorama. From these views, it is evident that these dense labels closely match the ground truth labels for objects such as roads, bushes, and buildings.

%From the above results, it is evident that noise exists in the dense labels, especially in small object categories. To mitigate the uncertainty introduced by erroneous dense labels, we propose our ScatterNet in Sec. \ref{section: ScatterNet}, which reduces this effect through multimodal consistency constraints.

% \begin{figure}[htbp]
%   \centering
%   \includegraphics[width=0.90\linewidth]{figures/figure_9.pdf} % 图片的文件名和路径
%   \caption{3D dense label precision for each class. (a) Results on SemanticKITTI \cite{behley2019semantickitti}; (b) Results on nuScenes \cite{caesar2020nuscenes}.} % 图片的标题
%   \label{fig:9} % 图片的标签，用于引用
% \end{figure}

\subsubsection{Application of Scatter Image Annotation}
\label{Sec:3.2.4}
\par
\begin{figure}[htbp]
  \centering
  \includegraphics[width=0.95\linewidth]{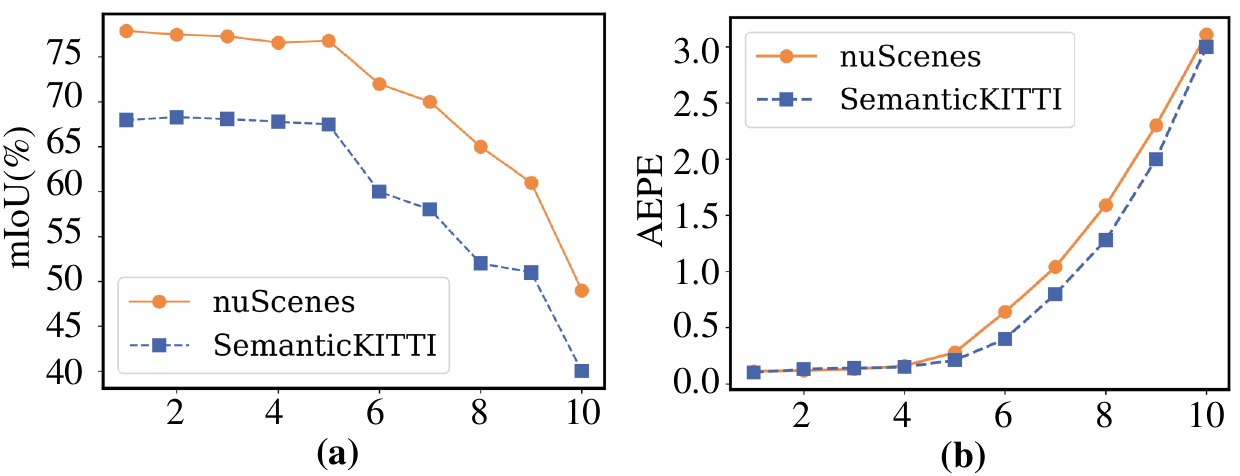} % 图片的文件名和路径
  \caption{mIoU and AEPE at different number of frames on the validation sets of SemanticKITTI \cite{behley2019semantickitti} and nuScenes \cite{caesar2020nuscenes}. (a) mIoU of dense labels. (b) AEPE of optical flow estimation.} % 图片的标题
  \label{fig:13} % 图片的标签，用于引用
\end{figure}

 Optical flow estimation errors increase with the number of frames used in interframe propagation. Therefore, determining the appropriate number of frames for interframe propagation is crucial for the application of the Scatter Image Annotation in other data collection scenarios. Our method determines the number of frames for sparse propagation within an acceptable mIoU error range by assessing the average end-point error (AEPE) of the optical flow estimation. On the validation sets of the SemanticKITTI and nuScenes datasets, we calculate the mIoU and AEPE for dense labels at different number of frames, as shown in Fig. \ref{fig:13}. The results indicate that when the AEPE is below 0.2 (i.e., fewer than 5 frames), the impact on the mIoU of dense labels is minimal. Thus, to account for a certain level of redundancy, we have set the number of propagation frames to 4. Extending this finding, the appropriate number of frames in other data collection scenarios can also be determined based on whether the AEPE is less than 0.2.

\subsection{ScatterNet}
\label{sec: 3.3}

\par We elaborately design a multimodal fusion framework called ScatterNet to reduce the performance gap caused by scatter image annotation. Specifically, ScatterNet comprises three components: (1) point-to-pixel correspondence, (2) multimodal fusion blocks, and (3) perceptual consistency loss. The overall scheme of ScatterNet is shown in Fig. \ref{fig:6}(b). To initiate the fusion process, we first establish the correspondence between point clouds and image pixels using perspective projection, which prepares for the fusion of images and point clouds. Then, we use an intermediate fusion branch to capture multimodal texture and structural features. Finally, we introduce a perceptual consistency loss to determine which information should be fused.

\subsubsection{Point to Pixel Correspondence}

\begin{figure}[!t]
  \centering
  \includegraphics[width=0.90\linewidth]{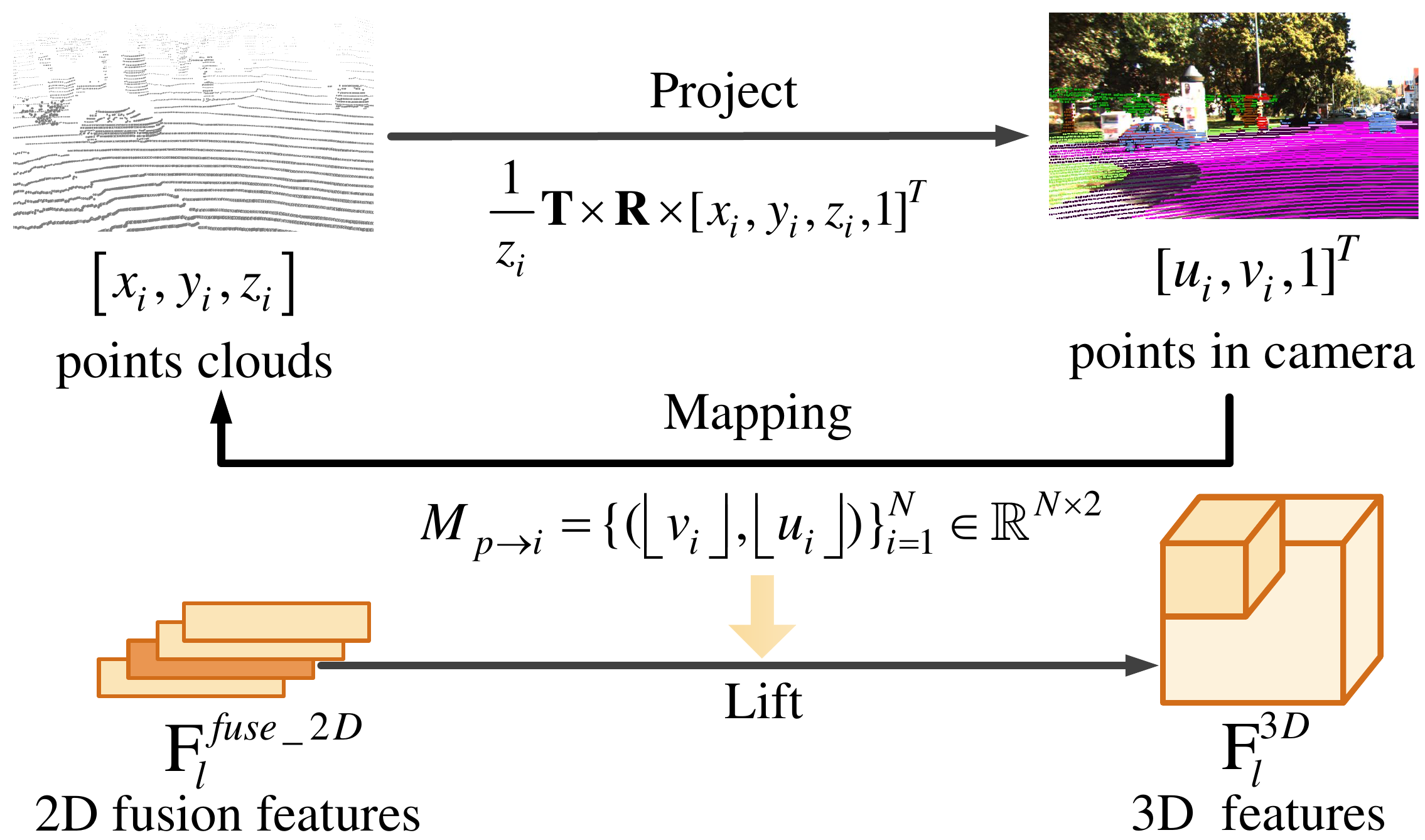} % 图片的文件名和路径
  \caption{Point to pixel correspondence. Using the camera's intrinsic and extrinsic parameters, the point cloud is projected onto the image plane, thereby establishing a mapping relationship between the point cloud and image pixels. This mapping enables 2D fusion features to be lifted to 3D features.} % 图片的标题
  \label{fig:14} % 图片的标签，用于引用
\end{figure}

Considering that \textit{Camera} and \textit{LiDAR} features are typically represented as pixels and points, transferring information directly between the two modalities is challenging. To integrate the features of the two modalities, we employ a two-stage fusion process based on the mapping relationship between points and pixels, as illustrated in Fig. \ref{fig:14}. 

\par \textit{In the first stage}, we follow \cite{zhuang2021perception} to project point clouds onto an image coordinate system. Specifically, consider $\mathbf{P}=\{{{p}_{i}}\}_{i=1}^{N}\in {{\mathbb{R}}^{N\times 3}}$ as the set of points. Each point ${{p}_{i}}=\left( {{x}_{i}},{{y}_{i}},{{z}_{i}} \right)\in {{\mathbb{R}}^{3}}$ is projected onto the image plane as a point ${{\hat{p}}_{i}}=({{u}_{i}},{{v}_{i}})\in {{\mathbb{R}}^{2}}$. The projection process is described by the following equation:
\begin{align}
\label{eq1}
{{({{u}_{i}},{{v}_{i}},1)}^{T}}=\frac{1}{{{z}_{i}}}\mathbf{T}\times \mathbf{R}\times {{({{x}_{i}},{{y}_{i}},{{z}_{i}},1)}^{T}},
\end{align}

where $\mathbf{T}\in {{\mathbb{R}}^{3\times 4}}$ and $\mathbf{R}\in {{\mathbb{R}}^{4\times 4}}$ are the intrinsic and extrinsic parameters of the camera, respectively, provided in the dataset.

Let ${{\mathbf{X}}_{img}}\in {{\mathbb{R}}^{3\times H\times W}}$ be the image, where $H$ and $W$ represent its height and width, respectively. Let ${{\mathbf{X}}_{point}}\in {{\mathbb{R}}^{N\times C}}$ be the LiDAR point cloud features, where $N$ and $C$ indicate the number of points and the dimension of the LiDAR features, respectively. Let ${{\mathbf{X}}_{proj}}\in {{\mathbb{R}}^{C\times H\times W}}$ be the projected point clouds obtained using Eq. (\ref{eq1}). Given the sparse nature of the point cloud, it is possible that certain pixels within ${{\mathbf{X}}_{proj}}$ lack a corresponding point from the set $\mathbf{P}$. To address this, we begin by setting all the pixel values to zero. Following \cite{zhuang2021perception}, we then generate 5-channel LiDAR projection features, i.e. $(d,x,y,z,i)$, for each pixel $(h,w)$ in ${{\mathbf{X}}_{proj}}$, where $d=\sqrt{{{x}^{2}}+{{y}^{2}}+{{z}^{2}}}$, $(x,y,z)$ represents the coordinates of the point cloud in the Cartesian coordinate system, and $i$  represents the intensity of the point cloud. The specific fusion details will be described in Sec. \ref{section:ScatterNet}.

\par After the projection, the point-to-pixel mapping is represented as:
\begin{equation}
\label{eq2}
{{M}_{p\to i}}=\{(\left\lfloor {{v}_{i}} \right\rfloor ,\left\lfloor {{u}_{i}} \right\rfloor) \}_{i=1}^{N}\in {{\mathbb{R}}^{N\times 2}},
\end{equation}

where $\left\lfloor \cdot  \right\rfloor $ is the floor operation.

\par \textit{In the second stage}, following Eq. (\ref{eq2}), we utilize the mapping relationship between points and pixels to lift the 2D fusion features to 3D space, thereby facilitating the fusion of 2D and 3D features. The detailed fusion process will be detailed in Sec. \ref{section:ScatterNet}.

\subsubsection{Multimodal Fusion Blocks}
\label{section:ScatterNet}

\begin{figure}[!t]
  \centering
  \includegraphics[width=0.90\linewidth]{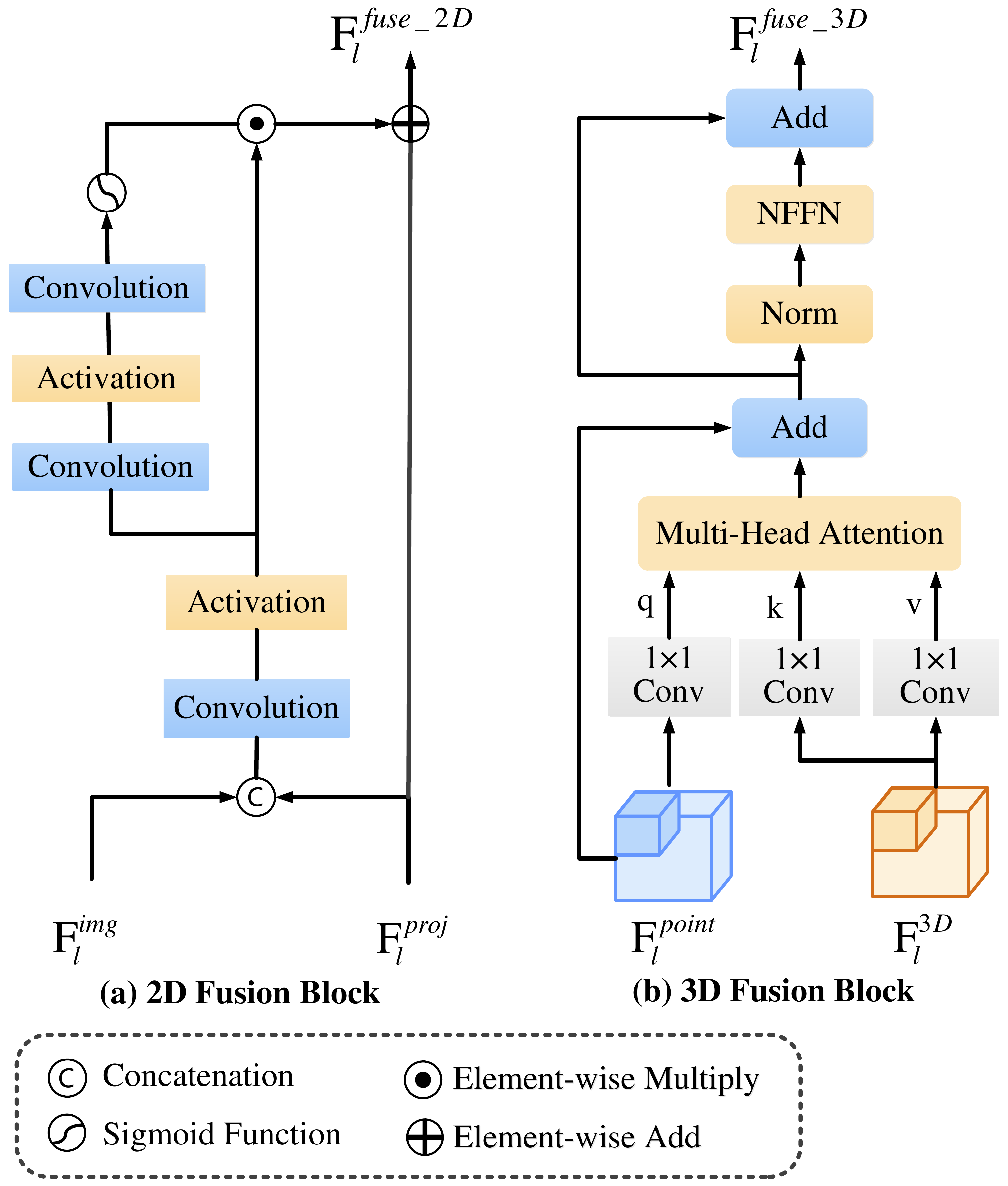} % 图片的文件名和路径
  \caption{Multimodal fusion blocks. (a) 2D fusion block: The projected point cloud and image features are fused through an attention mechanism.
(b) 3D fusion block: The cross attention mechanism is used to fuse point cloud features with 3D features.} % 图片的标题
  \label{fig:15} % 图片的标签，用于引用
\end{figure}

We propose an independent \textit{Fusion Stream} to integrate the structural and semantic features from the \textit{LiDAR} and \textit{Camera Streams}. Further details of the fusion process follow:

\par Let ${{M}_{img}}$, ${{M}_{poin}}$, and ${{M}_{fuse}}$ be the \textit{Camera}, \textit{LiDAR} and \textit{Fusion Streams}, respectively. Let ${{\mathbf{O}}^{img}}\in {{\mathbb{R}}^{S\times H\times W}}$, ${{\mathbf{O}}^{point}}\in {{\mathbb{R}}^{N\times S}}$ and ${{\mathbf{O}}^{fuse}}\in {{\mathbb{R}}^{S\times H\times W}}$ be the output probabilities for each stream, where $S$ indicates the number of semantic classes. It is important to note that each stream has four scales of output and classification heads. For clarity, we use one of these scales as an example. The outputs of these streams are computed by
\begin{equation}
\label{eq3a}
{{\mathbf{O}}^{img}}={{M}_{img}}({{\mathbf{X}}_{img}}),
\end{equation}
% \vspace{-4mm}
\begin{equation}
\label{eq3b}
{{\mathbf{O}}^{point}}={{M}_{point}}({{\mathbf{X}}_{point}}),
\end{equation}
% \vspace{-4mm}
\begin{equation}
\label{eq3c}
{{\mathbf{O}}^{fuse}}={{M}_{fuse}}({{\mathbf{X}}_{proj}}).
\end{equation}

\par For the \textbf{Camera stream}, following \cite{yan20222dpass} we crop the image into smaller patches as input data for the network. 
We use a ResNet34 \cite{he2016deep} encoder and a UNet \cite{ronneberger2015u} decoder to upscale the features of each encoder layer.

For the \textbf{LiDAR stream}, we adopt SPVCNN \cite{tang2020searching} to encode the features from the point clouds. For decoding, we use the multiscale point cloud upsampling method from 2DPASS \cite{yan20222dpass}.

\par For the \textbf{Fusion Stream}, we fuse the images and point clouds through two stages. In the first stage, we fuse the features of the image from \textit{Camera Stream} with the point clouds projected into the image plane. Let $\{\mathbf{F}_{l}^{img}\in {{\mathbb{R}}^{C_{l}^{img}\times H_{l}^{{}}\times W_{l}^{{}}}}\}_{l=L}^{L}$ represent the image features from the \textit{Camera Stream}, where $C$ is the number of channels, and $H_{l}$ and $W_{l}$ are the height and width of the feature maps for the $l$-th layer. Let $\{\mathbf{F}_{l}^{proj}\in {{\mathbb{R}}^{C_{l}^{proj}\times H_{l}^{{}}\times W_{l}^{{}}}}\}_{l=L}^{L}$ represent the projected point cloud features, where $C_{l}^{proj}$ is the number of channels in the $l$-th layer. We concatenate $\mathbf{F}_{l}^{img}$ and $\mathbf{F}_{l}^{proj}$, then apply a convolutional layer to reduce the number of channels, as illustrated in Eq. (\ref{eq4}).
\begin{align}
\label{eq4}
    \mathbf{F}_{l}^{2D}={{f}_{l}}([\mathbf{F}_{l}^{img};\mathbf{F}_{l}^{proj}]),
\end{align}

\par where $[\cdot ;\cdot ]$ denotes the concatenation operation, and ${{f}_{l}}(\cdot )$ is the convolution operation for the $l$-th fusion module.

\par Finally, we use residual modules and attention modules (see Fig. \ref{fig:15}(a)) to obtain the fusion features of the first stage by
\begin{equation}
\label{eq5}
\mathbf{F}_{l}^{fuse\_2D}=\mathbf{F}_{l}^{proj}+\sigma ({{g}_{l}}(F_{l}^{2D}))\odot F_{l}^{2D},
\end{equation}

where $\sigma (x)={1}/{(1+{{e}^{-x}})}$ denotes the sigmoid function, ${{g}_{l}}(\cdot )$ represents the convolution operation in the attention module w.r.t. the $l$-th layer and $\odot $ denotes element-wise multiplication.

In the second stage, let $\{\mathbf{F}_{l}^{point}\in \mathbb{R}_{l}^{N\times C_{l}^{point}}\}_{l=1}^{L}$ represent the features from the \textit{LiDAR Stream}, where $C_{l}^{point}$ denotes the number of channels in the $l$-th layer and $N$ indicates the number of point clouds. To fuse point cloud features $\mathbf{F}_{l}^{point}$ into the \textit{Fusion Stream}, we first lift the fused features $\mathbf{F}_{l}^{fuse\_2D}$ to 3D space based on the mapping relationship between point clouds and pixels in Eq. (\ref{eq2}), thereby obtaining the 3D features $\mathbf{F}_{l}^{3D}$. Then, using the cross-attention mechanism, we fuse $\mathbf{F}_{l}^{point}$ and $\mathbf{F}_{l}^{3D}$. The final fused features $\mathbf{F}_{l}^{fuse\_3D}$ are calculated as follows:

\begin{equation}
\label{eq6a}
\mathbf{F}_{l}^{fuse\_3D}={\textrm{MHA}}([\mathbf{F}_{l}^{3D};\mathbf{F}_{l}^{point}]) +  \mathbf{F}_{l}^{point},
\end{equation}
\begin{equation}
\label{eq6b}
\mathbf{F}_{l}^{fuse\_3D}=\mathbf{F}_{l}^{fuse\_3D}+\textrm{NFFN}(\mathbf{F}_{l}^{fuse\_3D}),
\end{equation}

where MHA represents \textit{Multi-Head Attention} \cite{vaswani2017attention}, and NFFN represents the \textit{normalizer layer} and the \textit{feed forward network}, as illustrated in Fig. \ref{fig:15}(b).

\subsubsection{Construction of Perceptual Consistency Loss}

\par The construction of a perceptual consistency loss is significant in our ScatterNet. To determine which information needs to be fused and which needs to be discarded during the fusion process, we extract strong confidence features from the \textit{Fusion Stream} and \textit{Camera Stream} into the \textit{LiDAR Stream} by perceiving the differences between different streams. Our perceptual consistency loss consists of two parts: consistency between \textit{Camera and Fusion Streams}, and consistency between \textit{LiDAR and Fusion Streams}, as shown in Fig. \ref{fig:6}(b). 

\par To measure the perceptual confidence of the predictions with respect to the \textit{Camera Stream}, \textit{LiDAR Stream} and \textit{Fusion Stream}, we first compute the three entropy maps $\mathbf{E}_{i}^{img}\in {{\mathbb{R}}^{H\times W}}$, $\mathbf{E}_{i}^{point}\in {{\mathbb{R}}^{H\times W}}$ and $\mathbf{E}_{i}^{fuse}\in {{\mathbb{R}}^{H\times W}}$ respectively by
\begin{equation}
\label{eq7a}
\mathbf{E}_{i}^{img} = -\frac{1}{\log S} \sum_{s=1}^{S} \mathbf{O}_{i,s}^{img} \log (\mathbf{O}_{i,s}^{img}),
\end{equation}
\begin{equation}
\label{eq7b}
\mathbf{E}_{i}^{point} = -\frac{1}{\log S} \sum_{s=1}^{S} \mathbf{O}_{i,s}^{point} \log (\mathbf{O}_{i,s}^{point}),
\end{equation}
\begin{equation}
\label{eq7c}
\mathbf{E}_{i}^{fuse} = -\frac{1}{\log S} \sum_{s=1}^{S} \mathbf{O}_{i,s}^{fuse} \log (\mathbf{O}_{i,s}^{fuse}).
\end{equation}

\par Following \cite{renyi1961measures}, we use $\log S$ to normalize the entropy to (0,1]. The perceptual confidence map ${{\mathbf{C}}_{img}}$, ${{\mathbf{C}}_{point}}$ and ${{\mathbf{C}}_{fuse}}$ with respect to the \textit{Camera}, \textit{LiDAR} and \textit{Fusion Streams} are computed by ${{\mathbf{C}}_{img}}=1-{{\mathbf{E}}_{img}}$, ${{\mathbf{C}}_{point}}=1-{{\mathbf{E}}_{point}}$ and ${{\mathbf{C}}_{fuse}}=1-{{\mathbf{E}}_{fuse}}$, respectively. Please be aware that not all information from the \textit{Camera} and \textit{Fusion Streams} is beneficial. Inspired by \cite{zhuang2021perception,jaritz2020xmuda}, predictions with lower confidence scores are more likely to be inaccurate. By combining confidence thresholds, we measure the importance of perceptual information from \textit{Camera} and \textit{Fusion Streams} by

\begin{equation}
\label{eq8a}
\Omega_{i}^{f\to p} = 
\begin{cases} 
\max(\mathbf{C}_{fuse} - \mathbf{C}_{point}), & \text{if } \mathbf{C}_{fuse} > \tau \\
0, & \text{otherwise}
\end{cases},
\end{equation}
\begin{equation}
\label{eq8b}
\Omega_{i}^{i\to p} = 
\begin{cases} 
\max(\mathbf{C}_{img} - \mathbf{C}_{fuse}), & \text{if } \mathbf{C}_{img} > \tau \\
0, & \text{otherwise}
\end{cases},
\end{equation}

where, $\tau $ indicates the confidence threshold.

Finally, we construct our perceptual consistency loss to extract information from a single-modal data (\textit{Camera Stream}) and a multimodal data (\textit{Fusion Stream}) to the target-modal data (\textit{LiDAR Stream}), as shown in Eq. (\ref{eq9a}) and Eq. (\ref{eq9b}):

\begin{equation}
\label{eq9a}
\mathcal{L}_{per}^{f\to i}=\frac{1}{N}\sum\limits_{i=1}^{N}{\Omega _{i}^{f\to i}{{D}_{KL}}({{\mathbf{O}}^{fuse}}||{{\mathrm{sg}[\mathbf{O}}^{img}]})},
\end{equation}

\begin{equation}
\label{eq9b}
\mathcal{L}_{per}^{p\to f}=\frac{1}{N}\sum\limits_{i=1}^{N}{\overset{{}}{\mathop{\Omega _{i}^{p\to f}}}\,}{{D}_{KL}}({{\mathbf{O}}^{point}}||{{\mathrm{sg}[\mathbf{O}}^{fuse}]}),
\end{equation}

where $N$ is the number of point clouds projected onto the image plane, `sg' stands for the stop gradient operator and ${{D}_{KL}}(\cdot ||\cdot )$indicates the Kullback-Leibler divergence \cite{hinton2015distilling}.

\par Apart from the perceptual consistency loss, we also use Cross entropy loss and Lov´asz-softmax loss \cite{berman2018lovasz} to train the network. These losses together form the loss function for the \textit{Camera, LiDAR, and Fusion Streams}, as defined below:

\begin{equation}
\label{eq10a}
{{\mathcal{L}}_{Camera}} = \mathcal{L}_{ce} + \mathcal{L}_{lov} + \lambda \mathcal{L}_{per}^{f\to i},
\end{equation}
\begin{equation}
\label{eq10b}
{{\mathcal{L}}_{Fusion}} = \mathcal{L}_{ce} + \mathcal{L}_{lov} + \gamma \mathcal{L}_{per}^{p\to f},
\end{equation}
\begin{equation}
\label{eq10c}
{{\mathcal{L}}_{Point}} = \mathcal{L}_{ce} + \mathcal{L}_{lov},
\end{equation}

where $\lambda$ and $\gamma$ are two hyperparameters that balance the various losses.

\section{Experiments}
In this part, we conduct experiments to provide a thorough assessment of our method.

\subsection{Experimental Setting}
\subsubsection{Datasets} 
Our method is evaluated on the nuScenes \cite{caesar2020nuscenes} and SemanticKITTI \cite{behley2019semantickitti} datasets. nuScenes is a multimodal autonomous driving dataset that includes six cameras and a 32-beam LiDAR sensor. It includes 1,000 driving scenes captured in Boston and Singapore, covering various weather and lighting conditions. SemanticKITTI, derived from the KITTI odometer benchmark \cite{geiger2013vision}, is a large-scale scan dataset captured in Germany. It offers 43,000 annotated scans in total, with sequences 00-10 (23,201 scans) divided into 19,130 for training and 4,071 for validation. In contrast to nuScenes, SemanticKITTI includes only front-view images.

\subsubsection{Cross-Modal Data Processing} 
Following \cite{sun2024image}, we filter out all points outside the camera field of view to establish the corresponding training and evaluation subsets. These subsets maintain a consistent point count in both the training and validation sets, ensuring a representative sampling of the entire data distribution. For instance, about 76.8\% of the data in nuScenes \cite{caesar2020nuscenes} is encompassed within these subsets. To ensure an equitable comparison, all comparative methods are trained using their original datasets and configurations, and predictions are collected only in the evaluation subset during assessment.

\subsubsection{Evaluation Protocols} 
Following \cite{sun2024image,qi2017pointnet}, we use the mIoU as the primary evaluation metric. In the weakly supervised setting, only dense 2D and 3D labels are used for training and validation, with true labels used for testing. In contrast, in the fully supervised setting, true labels are used for training, validation, and testing. Test-Time Augmentation is applied in both settings. Additionally, we provide the relative performance between Weakly Supervised training (WS) and the Fully Supervised upper bound (FS), denoted as WS/FS.

\subsection{Implementation Details}
We use SPVCNN \cite{tang2020searching} as the backbone for the \textit{LiDAR Stream}, while ResNet34 \cite{he2016deep} serves as the backbone for the \textit{Fusion and Camera Streams}, with features extracted after each downsampling layer to generate 2D features. The voxel size for SPVCNN \cite{tang2020searching} is set to 0.05 meters. The confidence threshold is set to 0.7. The initial learning rate is 0.24 and gradually reduces to 0 using a cosine annealing schedule. The batch size varies by dataset: 8 for SemanticKITTI and 4 for nuScenes. The parameters $\lambda$ and $\gamma$ are set to 0.2 and 0.5, respectively. In addition, we also apply a series of data augmentation techniques to prevent overfitting during training, including random cropping, instance augmentation, random horizontal flipping, and color jitter.

\subsection{Results on nuScenes}
\begin{table*}[t]
\centering
\caption{quantitative results of different approaches on nuScenes validation set}
\label{tab:1}
\setlength{\tabcolsep}{4.5pt}
\begin{tabular}{l|l|llllllllllllllll|ll}
\toprule
Methods    & Annot.    & \rotatebox{90}{barrier} & \rotatebox{90}{bicycle} & \rotatebox{90}{bus}  & \rotatebox{90}{car}  & \rotatebox{90}{construction} & \rotatebox{90}{motorcycle} & \rotatebox{90}{pedestrian} & \rotatebox{90}{traffic\_cone} & \rotatebox{90}{trailer} & \rotatebox{90}{truck} & \rotatebox{90}{driveable} & \rotatebox{90}{Other\_flat} & \rotatebox{90}{sidewalk} & \rotatebox{90}{terrain} & \rotatebox{90}{manmade} & \rotatebox{90}{vegetation} & \rotatebox{90}{mIoU(\%)}  & \rotatebox{90}{WS/FS(\%)} \\ 

\midrule
\specialrule{0em}{1.5pt}{1.5pt}
\midrule

SPVCNN \cite{tang2020searching}     & \multirow{9}{*}{100\%}  & 70.3    & 30.2    & 85.8 & 91.1 & 42.6         & 73.5       & 74.1       & 56.8         & 54.7    & 81.3  & 93.7      & 64.2       & 68.7     & 73.3    & 86.6    & 85.3       & 70.8  & -     \\
PolarNet \cite{zhang2020polarnet}    &        & 74.7    & 28.2    & 85.3 & 90.9 & 35.1         & 77.5       & 71.3       & 58.8         & 57.4    & 76.1  & 96.5      & 71.1     & 74.7     & 74.0      & 87.3    & 85.7       & 71.0    & -     \\
Salsanext \cite{cortinhal2020salsanext} &      & 74.8    & 34.1    & 85.9 & 88.4 & 42.4      & 72.4       & 72.2       & 63.1         & 61.3    & 76.5  & 96.0        & 70.8       & 71.2     & 71.5    & 86.7    & 84.4       & 72.2  & -     \\
Cylinder3D \cite{zhu2021cylindrical} &        & 74.5    & 43.1    & 87.4 & 85.9 & 45.1         & 80.2       & 79.7       & 65.3         & 61.5    & 80.6  & 96.5      & 71.2       & 74.9     & 75.3    & 87.7    & 87.1       & 74.8  & -     \\
AMVNet \cite{liong2020amvnet}    &    & 79.8    & 32.4    & 82.2 & 86.4 & 62.5         & 81.9       & 75.3       & 72.3         & 83.5    & 65.1  & 97.4      & 67.0         & 78.8     & 74.6    & 90.8    & 87.9       & 76.1  & -     \\
RPVNet \cite{xu2021rpvnet}    &       & 78.2    & 43.4    & 92.7 & 93.2 & 49.0         & 85.7       & 80.5       & 66.0       & 66.9    & 84.0    & 96.9      & 73.5       & 75.9     & 76.0    & 90.6    & 88.9       & 77.6  & -     \\
2DPASS \cite{yan20222dpass}    &       & 76.1    & 48.5    & 96.2 & 92.2 & 61.2         & 87.7       & 82.0       & 64.0         & 71.3    & 87.1  & 96.8      & 74.0       & 77.3     & 75.0    & 88.1    & 86.9       & 79.0    & -     \\
RangeViT \cite{ando2023rangevit}  &        & 75.5    & 40.7    & 88.3 & 90.1 & 49.3     & 79.3       & 77.2       & 66.3       & 65.2    & 80.0    & 96.4      & 71.4       & 73.8     & 73.8    & 89.9    & 87.2       & 75.2 & -     \\
Spherical \cite{lai2023spherical}     &    & 77.7    & 43.8    & 94.5 & 93.1 & 52.4         & 86.9       & 81.2       & 65.4         & 73.4    & 85.3  & 97.0      & 73.4       & 75.4     & 75.0    & 91.0    & 89.2       & 78.4  & -     \\ 

LIF-Seg \cite{10128757}     &    & 76.5    & 51.4    & 91.5 & 89.2 & 58.4  & 86.6       & 82.7       & 72.9         & 65.5    & 84.1  & 96.7     & 73.2       & 74.4     & 73.1    & 87.5    & 87.6       & 78.2  & -     \\ \midrule

PMF \cite{zhuang2021perception}  & \multirow{3}{*}{76.8\%} & 74.1    & 46.6    & 89.8 & 92.1 & 57.0         & 77.7       & 80.9       & 70.9         & 64.6    & 82.9  & 95.5      & 73.2       & 73.6     & 74.8    & 89.4    & 87.7       & 76.9  & -     \\
LUPC \cite{sun2024image}      &       & 75.2    & 56.5    & 91.7 & 92.9 & 62.8         & 81.7       & 82.7       & 70.7      & 64.5    & 83.3  & 95.4      & 74.7       & 73.8     & 75.4    & 90.0      & 88.6       & 78.8  & -     \\
\rowcolor{green!20}  % 设置灰色背景，20%的灰度
\textbf{Ours}        &      & 77.1    & 46.5    & 96.3 & 93.5 & 60.4         & 87.3       & 81.8       & 65.1       & 73.9    & 89.3  & 96.7    & 75.1   & 76.5     & 76.4    & 89.0      & 86.9       & \textbf{79.5}  & -     \\ \midrule
SLidR \cite{sautier2022image}     & 0.8\%         & 69.6    & 28.6    & 88.5 & 90   & 42.9         & 66.1       & 64.2       & 47.6         & 61.2    & 80.7  & 95.0      & 71.1       & 70.9     & 71.4    & 88.2    & 86.4       & 70.1  & 94.0    \\
Contra \cite{hou2021exploring}    & 0.2\%            & -       & -       & -    & -    & -            & -          & -          & -            & -       & -     & -         & -          & -        & -       & -       & -          & 63.5  & 84.2    \\
LESS \cite{liu2022less}      & 0.2\%      & -       & -       & -    & -    & -            & -          & -          & -         & -       & -     & -         & -          & -        & -       & -       & -          & 73.5  & 97.5     \\
IUPC \cite{sun2024image}      & 0.8\%      & 75.0    & 54.8    & 90.8 & 91.7 & 59.0       & 83.5       & 81.2       & 67.9      & 61.5    & 80.0  & 95.3      & 73.4       & 73.0       & 74.4    & 90.0    & 88.9       & 77.5  & \textbf{98.4}  \\
\rowcolor{green!40}  % 设置灰色背景，20%的灰度
\textbf{Ours}        & \textbf{0.02\%}                     & 79.3    & 35.2    & 90.7 & 88.5 & 73.5         & 80.2       & 80.4       & 75.1         & 84.4    & 74.2  & 93.9      & 65.0       & 77.5     & 73.2    & 89.4    & 86.2       & \textbf{77.9}  & 98.0    \\  \bottomrule 
\end{tabular}
\end{table*}

\par Table \ref{tab:1} presents a quantitative analysis of our method on the nuScenes dataset \cite{caesar2020nuscenes}. Despite our annotations covering only a tiny portion of the dataset (0.02\%), our method achieves performance comparable to almost all the latest fully supervised methods. In a fully supervised setting, our multimodal fusion method outperforms single-modal methods, such as the spherical method \cite{lai2023spherical}, by 1.1\% in mIoU. This result demonstrates that multimodal fusion can enhance the LiDAR features. Additionally, compared to previous multimodal methods such as 2DPASS \cite{yan20222dpass}, our method achieves a 0.5\% improvement in mIoU, confirming the effectiveness of our multimodal fusion strategy.

To further evaluate the weak supervision performance of our method, we compare it with other weakly supervised LiDAR segmentation methods, such as LESS \cite{liu2022less} and IUPC \cite{sun2024image}. The results indicate that despite a significant reduction in the number of annotations used, our performance remains comparable to existing methods. For example, compared to LUPC \cite{sun2024image}, our annotation strategy has reduced the amount of labeling by 97.5\%, while the performance has only decreased by 0.4\%. This performance improvement can be attributed primarily to two key factors. Firstly, adopting the scatter image annotation strategy effectively propagates manual annotations into dense labels, significantly reducing the cost of manual annotation. Secondly, ScatterNet significantly improves performance by integrating multimodal features through a dedicated fusion branch and using dense 2D labels for supervision to tackle modality imbalance issues. Moreover, it integrates a perceptual consistency loss function to mitigate the negative impact of low-confidence multimodal features on the network.

\subsection{Results on SemanticKITTI}
\begin{table*}[t!]
\centering
\caption{quantitative results of different approaches on SemanticKITTI validation set}
\label{tab:2}
\setlength{\tabcolsep}{3.0pt}
\begin{tabular}{l|l|lllllllllllllllllll|lll}
\toprule
Methods    & Annot.   & \rotatebox{90}{car}     & \rotatebox{90}{bicycle}   & \rotatebox{90}{motorcycle}    & \rotatebox{90}{truck}    & \rotatebox{90}{other-vehicle}    & \rotatebox{90}{Person}    & \rotatebox{90}{bicyclist}    & \rotatebox{90}{motorcyclist}     & \rotatebox{90}{road}     & \rotatebox{90}{parking}    & \rotatebox{90}{sidewalk}     & \rotatebox{90}{Other-ground}      & \rotatebox{90}{building}     & \rotatebox{90}{fence}     & \rotatebox{90}{vegetation}    & \rotatebox{90}{truck}     & \rotatebox{90}{terrian}    & \rotatebox{90}{pole}    & \rotatebox{90}{traffic-sign} & \rotatebox{90}{mIoU(\%)} & \rotatebox{90}{WS/FS(\%)} \\ 

\midrule
\specialrule{0em}{1.5pt}{1.5pt}
\midrule

SPVCNN \cite{tang2020searching}   & \multirow{5}{*}{100\%} & 97.1   & 35.2   & 64.6   & 72.7   & 64.3    & 69.7     & 82.5      & 0.2       & 93.5     & 50.8    & 81.0      & 0.3    & 91.1      & 63.5      & 89.2     & 66.1        & 77.2     & 64.1          & 49.4 & 63.8 & - \\
Cylinder3D  \cite{zhu2021cylindrical}   &  & 97.2        & 54.8    &78.9   & 79.6        & 67.1         & 75.4     & 86.2   & 1.0      & 94.7     & 45.6    & 81.7     & 0.6     & 89.3     & 53.0    & 87.0       & 70.8      & 71.4       & 65.7       & 52.8        & 65.9 & - \\
2DPASS \cite{yan20222dpass}  &     & 96.8     & 52.5      & 76.3        & 90.7     & 71.3       & 78.3       & 92.3      & 0.1       & 93.2      & 50.7       & 80.1      & 8.4     & 92.2      & 68.2     & 88.4    & 71.2        & 74.6    & 63.9       & 53.5 & 68.6 & - \\
RangeViT \cite{ando2023rangevit} &   & 95.4 & 55.8 & 43.5 & 29.8 & 42.1 & 63.9 & 58.2 & 38.1 & 93.1 & 70.2 & 80.0 & 32.5 & 92.0 & 69.0 & 85.3 & 70.6 & 71.2 & 60.8 & 64.7 & 64.0 & - \\ \midrule

PMF \cite{zhuang2021perception}   & \multirow{3}{*}{15.9\%}  & 95.4       & 47.8     & 62.9      & 68.4     & 75.2      & 78.9      & 71.6        & 0.0     & 96.4       & 43.5    & 80.5         & 0.1     & 88.7    & 60.1    & 88.6     & 72.7       & 75.3      & 65.5     & 43.0 & 63.9 & -- \\
IUPC \cite{sun2024image}     &        & 97.0        & 43.0    & 62.1      & 76.7       & 71.5    & 84.7      & 89.9       & 0.0     & 95.6      & 38.5       & 79.3       & 0.2         & 89.3       & 59.4         & 88.4   & 73.6      & 73.3         & 67.2         & 43.8 & 64.9 & - \\
\rowcolor{green!20}  % 设置灰色背景，20%的灰度
\textbf{Ours}        &        & 97.3      & 52.2     & 64.6      & 91.2         & 73.2        & 79.2      & 92.1         & 7.4        & 94.5       & 45.6          & 82.5          & 11.6         & 92.1       & 70.4      & 88.3     & 72.6     & 73.7      & 64.8      & 60.9 & \textbf{69.2} & - \\ \midrule
Scribble \cite{unal2022scribble}      & 8.0\%     & 91.2      & 42.1         & 64.8       & 80.7      & 65.6      & 81.4     & 88.5        & 0.0        & 90.0        & 32.1       & 70.6          & 3.1       & 88.2       & 53.0      & 86.3        & 68.6      & 69.5        & 61.5         & 43.3 & 61.9 & 96.3 \\
ReDAL \cite{wu2021redal}     & 5.0\%         & 95.4      & 29.6      & 58.6         & 63.4         & 49.8        & 63.4         & 84.1      & 0.5          & 91.5        & 39.3       & 78.4    & 1.2        & 89.3      & 54.4       & 87.4         & 62.0       & 74.1      & 63.5       & 49.7 & 59.8 & 97.4 \\

OTOC \cite{liu2021one} & 0.1\% &  77.0  & 0.0     & 0.0     & 2.0     & 1.0         & 0.0    & 2.0      & 0.0      & 63.0     & 0.0      & 38.0     & 0.0   & 73.0         & 44.0       & 78.0        & 39.0        & 53.0        & 25.0      & 0.0 & 26.0 & 39.4 \\

HybridCR \cite{li2022hybridcr}      & 1.0\%      & -     & -      & -      & -     & -     & -        & -      & -       & -     & -     & -      & -      & -    & -          & -       & -       & -        & -      & - & 51.9 & 97.6 \\
SQN \cite{hu2022sqn}   & 0.10\%         & 92.1       & 39.3          & 30.1         & 36.7       & 26.0         & 36.4       & 25.3      & 7.2     & 90.5    & 56.8        & 72.9      & 19.1      & 84.8     & 53.3       & 80.8     & 59.1     & 67.0     & 44.5     & 44.0  & 50.8 & 95.5 \\
SLidR  \cite{sautier2022image}        & 0.08\%     & 94.1     & 5.8    & 26.8       & 70.9         & 57.1      & 62.0    & 81.6    & 0.0      & 94.7       & 34.6        & 75.1            & 0.2     & 85.3      & 40.3     & 83.1     & 55.9    & 66.1     & 49.0    & 25.7 & 53.1 & 93.2 \\
Contra \cite{hou2021exploring}  & 0.1\% &  93.0  & 0.0     & 0.0     & 62.0     & 45.0         & 28.0    & 0.0      & 0.0      & 90.0     & 39.0      & 71.0     & 6.0   & 90.0         & 42.0       & 89.0        & 57.0        & 75.0        & 54.0      & 34.0 & 46.0 & 69.8 \\

LESS \cite{liu2022less}      & 0.01\%    & 96.0     & 33.0     & 61.0     & 73.0     & 59.0         & 68.0    & 87.0      & 0.0      & 92.0     & 38.0      & 76.0     & 5.0   & 89.0         & 52.0       & 87.0        & 67.0        & 71.0        & 59.0      & 46.0  & 61.0 & 92.5 \\
OPOCA \cite{huang2024opoca}    & 0.092\%    & 85.5   & 23.3    & 53.0    & 66.2      & 35.4        & 39.0    & 62.3     & 0.0    & 91.3     & 36.4    & 78.0    & 4.8       & 84.8     & 37.3       & 82.7       & 49.4     & 74.3    & 28.4      & 25.8 & 50.4 & 85.2 \\
IUPC  \cite{sun2024image}     & 0.08\%     & 95.6    & 48.7   & 63.5    & 75.6     & 67.3       & 75.3     & 79.8     & 0.0    & 95.0     & 38.9       & 78.2           & 2.0      & 89.9    & 62.1    & 88.2    & 71.6    & 71.9      & 62.7    & 45.1 & 63.7 & \textbf{98.2} \\
\rowcolor{green!40}  % 设置灰色背景，20%的灰度
\textbf{Ours}   & \textbf{0.004\%}     & 93.2        & 51.0     & 58.6     & 57.3       & 58.9       & 75.5     & 75.0      & 59.5      & 84.6     & 61.5      & 67.5     & 33.8        & 88.1         & 63.0      & 82.2       & 68.2         & 66.7     & 58.7      & 45.1 & \textbf{65.7} & 95.0 \\ \bottomrule
\end{tabular}
\end{table*}

\par In Table \ref{tab:2}, we present the quantitative results for the SemanticKITTI dataset \cite{behley2019semantickitti}. It is important to note that this dataset only includes front-view images. Therefore, we follow the method in \cite{sun2024image} and test using only the front-view LiDAR.

\par To evaluate the weak supervision performance of our method, we compare it with existing weakly supervised methods. These methods include Contra \cite{hou2021exploring}, ReDAL \cite{wu2021redal}, OTOC \cite{huang2024opoca}, and SQN \cite{hu2022sqn}, which are initially tested only on indoor datasets. Therefore, we use the reproduced results from \cite{liu2022less} for comparison. Since SLidR \cite{sautier2022image} is not specifically designed for weakly supervised LiDAR segmentation, we use the experimental settings and results from \cite{sun2024image} for a fair comparison. Furthermore, we compare our method with approaches designed for outdoor scenarios, such as Scribble \cite{unal2022scribble}, LESS \cite{liu2022less}, and IUPC \cite{sun2024image}. Consistent with the conclusions on the nuscenss dataset: our method significantly reduces the annotation cost, while the decline in WS/FS performance is minimal. For example, compared to LUPC \cite{sun2024image}, our annotation strategy has reduced the amount of labeling by 95.0\%, while the performance has only decreased by 4.2\%.

 \begin{table}[]
 \centering
\caption{Effect of Network components on the SemanticKITTI
validation set. \textbf{PCL} indicates Perceptual consistency loss}
\label{tab:3}
\begin{tabular}{c|ccc|c}
\toprule
\multirow{2}{*}{LiDAR Stream} & \multicolumn{3}{c|}{ScatterNet}    & \multicolumn{1}{l}{\multirow{2}{*}{mIoU (\%)}} \\
& \multicolumn{1}{l}{Camera Stream} & \multicolumn{1}{l}{Fusion Stream} & \multicolumn{1}{l|}{\textbf{PCL}} & \multicolumn{1}{l}{}      \\ \midrule
\checkmark    &   &      &  & 61.6   \\
\checkmark    & \checkmark  &   &                             & 62.1   \\
\checkmark   & \checkmark  & \checkmark  &  & 63.6   \\
\checkmark & \checkmark    & \checkmark                             & \checkmark     & 65.7   \\ \bottomrule
\end{tabular}
\end{table}
\vspace{1em} 

\begin{table}[htbp]
\centering
\begin{minipage}[t]{0.35\linewidth}
\centering
\caption{Comparing with different Consistency loss.}
\label{tab:4}
\begin{tabular}{l|c}
\toprule
Methods & mIoU \\ \midrule
Hinton et.al \cite{hinton2015distilling} & 63.5 \\
xMUDA \cite{jaritz2020xmuda} & 64.0 \\
PMF \cite{zhuang2021perception} & 63.9 \\
2DPASS \cite{yan20222dpass} & 64.2 \\
\rowcolor{green!40}  % 设置灰色背景，20%的灰度
\textbf{Ours} & \textbf{65.7} \\ \bottomrule
\end{tabular}
\end{minipage}%
\hfill
\begin{minipage}[t]{0.6\linewidth}
\centering
\caption{Comparing with other multimodal fusion structures.}
\label{tab:5}
\begin{tabular}{l|cc}
\toprule
Methods & mIoU(3D) & mIoU(2D) \\ \midrule
Baseline & 63.8 & 47.8 \\
xMUDA \cite{jaritz2020xmuda} & 64.2 & 48.8 \\
PMF \cite{zhuang2021perception} & 65.8 & 48.6 \\
2DPASS \cite{yan20222dpass} & 66.4 & 49.1 \\
\rowcolor{green!20}  % 设置灰色背景，20%的灰度
\textbf{Ours (sparse)} & \textbf{67.8} & \textbf{51.4} \\ 
\rowcolor{green!40}  % 设置灰色背景，20%的灰度
\textbf{Ours (dense)} & \textbf{69.2} & \textbf{59.4} \\ 
\bottomrule
\end{tabular}
\end{minipage}
\end{table}

\subsection{Ablation Study}
To demonstrate the contributions of different components in our ScatterNet, we conduct ablation experiments on SemanticKITTI \cite{behley2019semantickitti}.

\subsubsection{Effect of Network Components}
\par
\begin{figure*}[ht]
\centering
\setlength{\belowcaptionskip}{-3mm}
\includegraphics[width=0.95\linewidth]{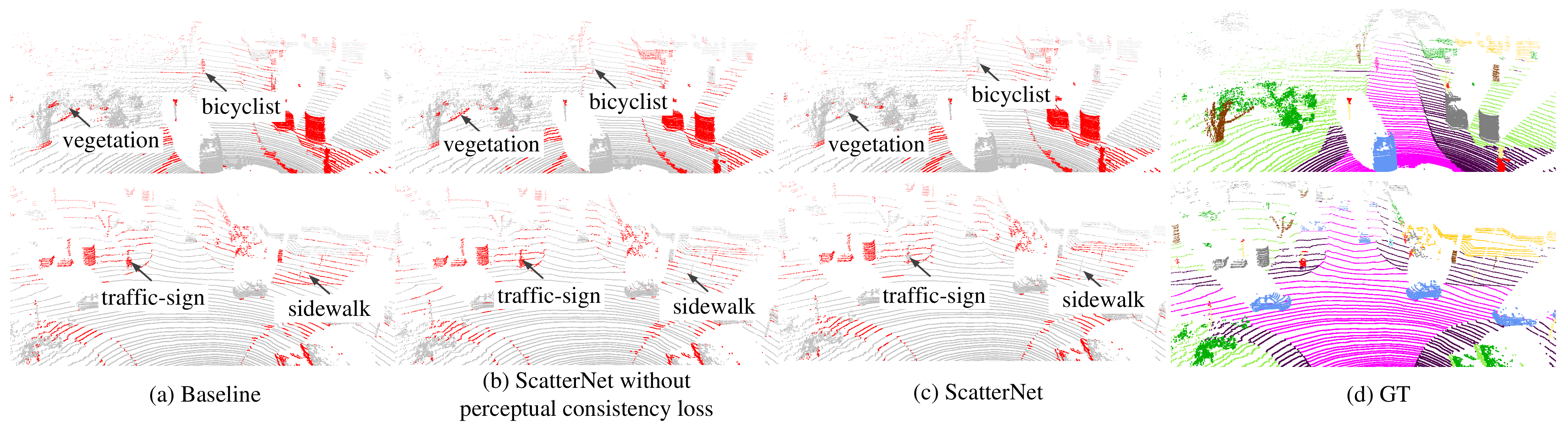}
\vspace{-0.0cm}
\caption{\textbf{Effect of ScatterNet Components.} (a) Errors using only the LiDAR stream, (b) Errors by ScatterNet without perceptual consistency loss, (c) Errors by ScatterNet, (d) Ground truth.}
\label{fig:16}
\end{figure*}

 The results are displayed in Table \ref{tab:3}. Compared to using only the \textit{LiDAR Stream}, adding the \textit{Camera Stream} results in a slight improvement of about 0.5\%, indicating that the \textit{Camera Stream} alone does not provide effective multimodal features for the \textit{LiDAR Stream}. However, introducing an independent \textit{Fusion Stream} significantly enhances segmentation performance, achieving an mIoU of 63.6\%, an increase of approximately 1.5\%. Furthermore, applying the loss of perceptual consistency to extract high-confidence multimodal features further boosts performance by approximately 2.1\%. These results demonstrate that our independent \textit{Fusion Stream} and perceptual consistency loss are key to enhancing performance. The related visualization results are shown in Fig. \ref{fig:16}.

\subsubsection{Comparing Perceptual Consistency Loss with Other Methods} 
\par To further verify the effectiveness of perceptual consistency loss, we compare it with other methods in Table \ref{tab:4} by replacing it in ScatterNet. We chose the pure-knowledge distillation loss \cite{hinton2015distilling} proposed by Hinton, a pioneer in this research field. In addition, we also select methods proposed in recent years, such as xMUDA \cite{jaritz2020xmuda}, PMF \cite{zhuang2021perception}, and 2DPASS \cite{yan20222dpass}. As shown in Table \ref{tab:4}, previous methods have certain limitations in the application of LiDAR semantic segmentation. For instance, the pure knowledge extraction method \cite{hinton2015distilling} has limited effectiveness in extracting multimodal knowledge; xMUDA \cite{jaritz2020xmuda} employs cross-modal feature alignment for domain adaptation, but it shows marginal performance improvement in LiDAR segmentation; PMF \cite{zhuang2021perception} uses a bidirectional perceptual loss, which complicates the decoupling of multimodal features from single-modal features; 2DPASS \cite{yan20222dpass} proposes multi-scale fusion for single knowledge utilization, yet lacks effective inter-modal information perception, resulting in low confidence in fusion information. In contrast, our method utilizes perceptual consistency loss, effectively helping the \textit{LiDAR Stream} to capture perceptual information from images and integrate features. This approach not only enhances the integration of inter-modal information but also significantly improves the precision and reliability of semantic segmentation.

\subsubsection{Comparing Multimodal Fusion Structures  with Other Methods}

\par
\begin{figure}[ht]
\centering
\setlength{\belowcaptionskip}{-3mm}
\includegraphics[width=0.90\linewidth]{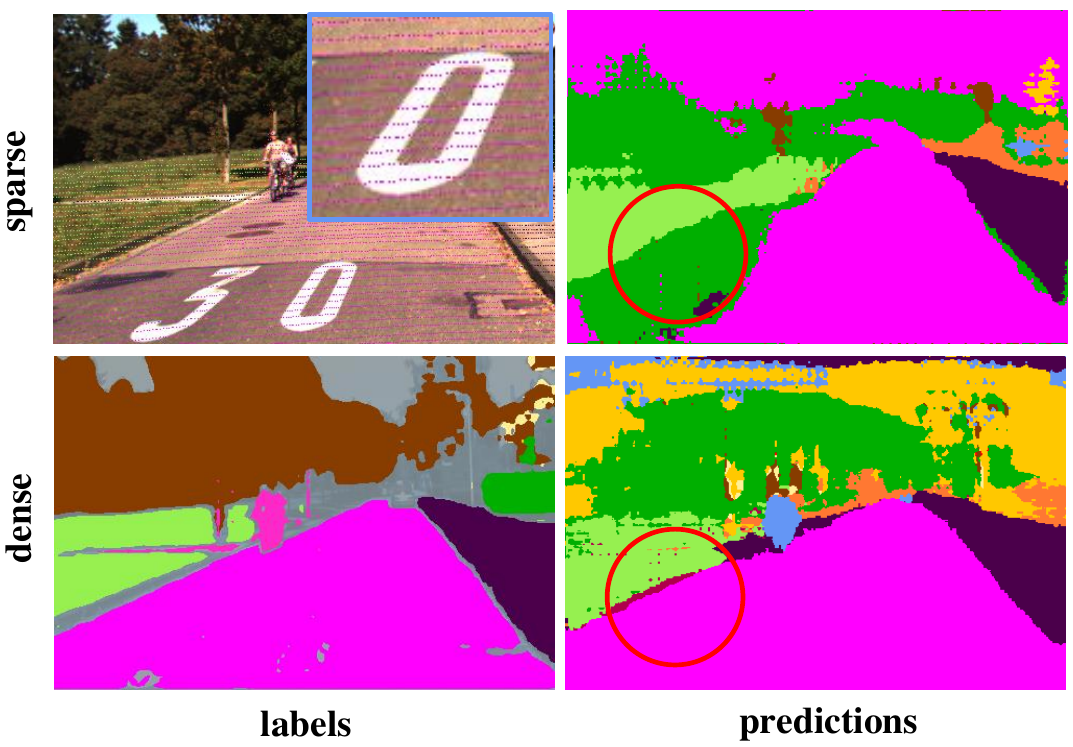}
\vspace{-0.0cm}
\caption{\textbf{Comparison of segmentation results from the camera stream.} The top shows the results using sparse 2D labels, while the bottom shows results using dense 2D labels. It is evident that the segmentation of road and terrain is significantly better with dense labels compared to sparse labels. Red circles highlight the differences between the results, and a blue box provides zoom-ins of a specific area.
}
\label{fig:17}
\end{figure}

 To further validate the advantages of our fusion scheme, we compare it against currently popular multimodal fusion frameworks. The baseline model is trained using only a single modality. To ensure a fair comparison, we use a unified backbone for feature extraction. Specifically, the \textit{LiDAR Stream} employs SPVCNN \cite{tang2020searching}, while the \textit{Camera Stream} utilizes UNet \cite{ronneberger2015u}. Additionally, considering the impact of the consistency loss function, all methods employ basic KL divergence. As seen in Table \ref{tab:5}, due to the lack of feature-level fusion between point clouds and images, xMUDA \cite{jaritz2020xmuda} shows limited performance improvement over the baseline (single modality). PMF \cite{zhuang2021perception} projects point clouds onto the image plane, achieving multiscale feature fusion during the feature encoding stage. However, this method discards the structural information of the point clouds. 2DPASS \cite{yan20222dpass} lifts image features to 3D space during the decoding stage. Although this method retains the structural information of the point cloud, not all image pixels correspond to the point cloud, resulting in a compromise of the texture features of images. Our method retains the advantages of previous methods and conducts feature fusion during both encoding and decoding stages through an intermediate fusion branch, thereby obtaining richer structural and texture features. Furthermore, if dense labels rather than traditional sparse labels are used for supervision in the \textit{Camera Stream}, our performance can be further improved by 1.4\%.  This enhancement primarily arises from the dense labels providing richer supervisory information, resulting in an increase in the mIoU from 59.4\% to 61.4\%. Additionally, we have visualized the outcomes of the \textit{Camera Stream}, as shown in Fig. \ref{fig:17}, demonstrating that dense 2D labels yield more accurate segmentation results.

\subsubsection{Generality}

\begin{table}[]
\centering
\caption{Generalization of ScatterNet with Different LiDAR Stream on SemanticKITTI Dataset.}
\label{tab:6}
\begin{tabular}{l|c|c|cc}
\toprule
Model                         & Supervision & Our & mIoU(\%) & WS/FS(\%) \\  
\midrule
\multirow{4}{*}{MinkowskNet \cite{choy20194d}} & Fully       &     & 64.4 & -     \\
                             & \cellcolor{green!20}Fully       & \cellcolor{green!20}\checkmark    & \cellcolor{green!20}65.9 & \cellcolor{green!20}-     \\
                             & Weakly      &     & 57.3 & 88.9    \\
                             & \cellcolor{green!40}Weakly      & \cellcolor{green!40}\checkmark    & \cellcolor{green!40}62.8 & \cellcolor{green!40}95.3    \\ 
\midrule
\multirow{4}{*}{Cylinder3D \cite{zhu2021cylindrical}}  & Fully       &     & 65.9 & -     \\
                             & \cellcolor{green!20}Fully       & \cellcolor{green!20}\checkmark    & \cellcolor{green!20}66.7 & \cellcolor{green!20}-     \\
                             & Weakly      &     & 58.3 & 88.5    \\
                             & \cellcolor{green!40}Weakly      & \cellcolor{green!40}\checkmark    & \cellcolor{green!40}63.3 & \cellcolor{green!40}95.0    \\ 
\midrule
\multirow{4}{*}{Spherical \cite{lai2023spherical}}   & Fully       &     & 68.2 & -     \\
                             & \cellcolor{green!20}Fully       & \cellcolor{green!20}\checkmark    & \cellcolor{green!20}69.4 & \cellcolor{green!20}-     \\
                             & Weakly      &     & 62.0 & 91.0    \\
                             & \cellcolor{green!40}Weakly      & \cellcolor{green!40}\checkmark    & \cellcolor{green!40}66.2 & \cellcolor{green!40}95.4    \\ 
\bottomrule
\end{tabular}
\end{table}

\par To validate the generalization of ScatterNet, we replace the original SPVCNN \cite{tang2020searching} in the \textit{LiDAR Stream} with three mainstream networks: MinkowskiNet \cite{choy20194d}, Cylinder3D \cite{zhu2021cylindrical}, and Spherical Transformer \cite{lai2023spherical}. Table \ref{tab:6} shows the LiDAR segmentation results on SemanticKITTI \cite{behley2019semantickitti}. Specifically, under fully supervised conditions, our multimodal fusion framework significantly improves the performance of the original single-modal point cloud segmentation networks. Additionally, in the weakly supervised setting, by adopting our method, WS/FS of the three networks improve from 88.9\%, 88.5\%, and 91.0\% to 95.3\%, 95.0\%, and 95.4\%, respectively. Overall, these results demonstrate that our approach can be effectively applied to various single-modal LiDAR semantic segmentation models and enhance their performance under both fully supervised and weakly supervised settings.

\subsubsection{Effect of hyperparameters \texorpdfstring{$\lambda$}{lambda} and \texorpdfstring{$\gamma$}{gamma}}

To investigate the impact of \texorpdfstring{$\lambda$}{lambda}, we first set \texorpdfstring{$\gamma$}{gamma} to 1 and train ScatterNet with $\lambda \in \left\{ 0.0,0.2,0.5,0.7,0.9 \right\}$ on SemanticKITTI \cite{behley2019semantickitti}. According to Table \ref{tab:7}, ScatterNet with $\lambda$ = 0.2 performs the best. Then, we keep \texorpdfstring{$\lambda$}{lambda} at 0.2 to train models with $\gamma \in \left\{ 0.0,0.2,0.5,1.0,2.0 \right\}$. It is evident that the ScatterNet achieves the best performance when \texorpdfstring{$\gamma$}{gamma} = 0.5. Therefore, in our experiments, we set \texorpdfstring{$\lambda$}{lambda}, \texorpdfstring{$\gamma$}{gamma} to 0.2 and 0.5, respectively.

\begin{table}[h]
\centering
\caption{Effects of $\lambda$ and $\gamma$. Best results are highlighted in bold.}
\label{tab:7}
\begin{tabular}{c|p{0.3cm}p{0.2cm}p{0.3cm}p{0.3cm}p{0.4cm}|p{0.3cm}p{0.3cm}p{0.3cm}p{0.3cm}p{0.3cm}}
\toprule
Parameter & \multicolumn{5}{c|}{$\lambda$} & \multicolumn{5}{c}{$\gamma$} \\
\midrule
Value     & 0.0  & 0.2    & 0.5    & 0.7    & 0.9  & 0.0  & 0.2  & 0.5  & 1.0  & 2.0  \\
\midrule
mIoU (\%) & 64.2 & \textbf{65.4} & 65.2 & 65.2 & 65.0 & 63.8 & 65.4 & \textbf{65.7} & 65.4 & 65.4 \\
\bottomrule
\end{tabular}
\end{table}

\section{Conclusion}
We have proposed a weakly supervised LiDAR semantic segmentation method that significantly reduces the dependency on extensive labeled data. Unlike previous point-based weak annotation strategies, our Image Scatter Annotation is a pixel-based approach. Additionally, we have developed a multimodal framework named ScatterNet, which adopts three key strategies: using dense semantic labels to supervise the image branch, proposing an intermediate fusion branch to enhance multimodal features, and introducing a perception consistency loss to mitigate the adverse effects of low-confidence multimodal features on the network. However, a limitation of our method is that it requires LiDAR and camera to be aligned both temporally and spatially, which can be challenging when images are unavailable or not aligned with the LiDAR.

% Can use something like this to put references on a page
% by themselves when using endfloat and the captionsoff option.
% \ifCLASSOPTIONcaptionsoff
%   \newpage
% \fi

% trigger a \newpage just before the given reference
% number - used to balance the columns on the last page
% adjust value as needed - may need to be readjusted if
% the document is modified later
%\IEEEtriggeratref{8}
% The "triggered" command can be changed if desired:
%\IEEEtriggercmd{\enlargethispage{-5in}}

% references section

% can use a bibliography generated by BibTeX as a .bbl file
% BibTeX documentation can be easily obtained at:
% http://mirror.ctan.org/biblio/bibtex/contrib/doc/
% The IEEEtran BibTeX style support page is at:
% http://www.michaelshell.org/tex/ieeetran/bibtex/
%\bibliographystyle{IEEEtran}
% argument is your BibTeX string definitions and bibliography database(s)
%\bibliography{IEEEabrv,../bib/paper}
%
% <OR> manually copy in the resultant .bbl file
% set second argument of \begin to the number of references
% (used to reserve space for the reference number labels box)
\bibliographystyle{IEEEtran}
\bibliography{references}

% Generated by IEEEtran.bst, version: 1.14 (2015/08/26)
\begin{thebibliography}{10}
\providecommand{\url}[1]{#1}
\csname url@samestyle\endcsname
\providecommand{\newblock}{\relax}
\providecommand{\bibinfo}[2]{#2}
\providecommand{\BIBentrySTDinterwordspacing}{\spaceskip=0pt\relax}
\providecommand{\BIBentryALTinterwordstretchfactor}{4}
\providecommand{\BIBentryALTinterwordspacing}{\spaceskip=\fontdimen2\font plus
\BIBentryALTinterwordstretchfactor\fontdimen3\font minus \fontdimen4\font\relax}
\providecommand{\BIBforeignlanguage}[2]{{%
\expandafter\ifx\csname l@#1\endcsname\relax
\typeout{** WARNING: IEEEtran.bst: No hyphenation pattern has been}%
\typeout{** loaded for the language `#1'. Using the pattern for}%
\typeout{** the default language instead.}%
\else
\language=\csname l@#1\endcsname
\fi
#2}}
\providecommand{\BIBdecl}{\relax}
\BIBdecl

\bibitem{10487013}
A.~Du, T.~Zhou, S.~Pang, Q.~Wu, and J.~Zhang, ``Pcl: Point contrast and labeling for weakly supervised point cloud semantic segmentation,'' \emph{IEEE Transactions on Multimedia}, pp. 1--12, 2024.

\bibitem{liu2021one}
Z.~Liu, X.~Qi, and C.-W. Fu, ``One thing one click: A self-training approach for weakly supervised 3d semantic segmentation,'' in \emph{Proceedings of the IEEE/CVF Conference on Computer Vision and Pattern Recognition}, 2021, pp. 1726--1736.

\bibitem{yang2022mil}
C.-K. Yang, J.-J. Wu, K.-S. Chen, Y.-Y. Chuang, and Y.-Y. Lin, ``An mil-derived transformer for weakly supervised point cloud segmentation,'' in \emph{Proceedings of the IEEE/CVF conference on computer vision and pattern recognition}, 2022, pp. 11\,830--11\,839.

\bibitem{zhang2021weakly}
Y.~Zhang, Z.~Li, Y.~Xie, Y.~Qu, C.~Li, and T.~Mei, ``Weakly supervised semantic segmentation for large-scale point cloud,'' in \emph{Proceedings of the AAAI Conference on Artificial Intelligence}, vol.~35, no.~4, 2021, pp. 3421--3429.

\bibitem{wei2020multi}
J.~Wei, G.~Lin, K.-H. Yap, T.-Y. Hung, and L.~Xie, ``Multi-path region mining for weakly supervised 3d semantic segmentation on point clouds,'' in \emph{Proceedings of the IEEE/CVF conference on computer vision and pattern recognition}, 2020, pp. 4384--4393.

\bibitem{li2024multi}
X.~Li, Q.~Xu, J.~Zhang, T.~Zhang, Q.~Yu, L.~Sheng, and D.~Xu, ``Multi-modality affinity inference for weakly supervised 3d semantic segmentation,'' in \emph{Proceedings of the AAAI Conference on Artificial Intelligence}, vol.~38, no.~4, 2024, pp. 3216--3224.

\bibitem{hu2022sqn}
Q.~Hu, B.~Yang, G.~Fang, Y.~Guo, A.~Leonardis, N.~Trigoni, and A.~Markham, ``Sqn: Weakly-supervised semantic segmentation of large-scale 3d point clouds,'' in \emph{European Conference on Computer Vision}.\hskip 1em plus 0.5em minus 0.4em\relax Springer, 2022, pp. 600--619.

\bibitem{liu2022less}
M.~Liu, Y.~Zhou, C.~R. Qi, B.~Gong, H.~Su, and D.~Anguelov, ``Less: Label-efficient semantic segmentation for lidar point clouds,'' in \emph{European conference on computer vision}.\hskip 1em plus 0.5em minus 0.4em\relax Springer, 2022, pp. 70--89.

\bibitem{liu2023cpcm}
L.~Liu, Z.~Zhuang, S.~Huang, X.~Xiao, T.~Xiang, C.~Chen, J.~Wang, and M.~Tan, ``Cpcm: Contextual point cloud modeling for weakly-supervised point cloud semantic segmentation,'' in \emph{Proceedings of the IEEE/CVF International Conference on Computer Vision}, 2023, pp. 18\,413--18\,422.

\bibitem{wang2020weakly}
H.~Wang, X.~Rong, L.~Yang, J.~Feng, J.~Xiao, and Y.~Tian, ``Weakly supervised semantic segmentation in 3d graph-structured point clouds of wild scenes,'' \emph{arXiv preprint arXiv:2004.12498}, 2020.

\bibitem{sautier2022image}
C.~Sautier, G.~Puy, S.~Gidaris, A.~Boulch, A.~Bursuc, and R.~Marlet, ``Image-to-lidar self-supervised distillation for autonomous driving data,'' in \emph{Proceedings of the IEEE/CVF Conference on Computer Vision and Pattern Recognition}, 2022, pp. 9891--9901.

\bibitem{sun2024image}
T.~Sun, Z.~Zhang, X.~Tan, Y.~Qu, and Y.~Xie, ``Image understands point cloud: Weakly supervised 3d semantic segmentation via association learning,'' \emph{IEEE Transactions on Image Processing}, 2024.

\bibitem{hou2021exploring}
J.~Hou, B.~Graham, M.~Nie{\ss}ner, and S.~Xie, ``Exploring data-efficient 3d scene understanding with contrastive scene contexts,'' in \emph{Proceedings of the IEEE/CVF Conference on Computer Vision and Pattern Recognition}, 2021, pp. 15\,587--15\,597.

\bibitem{behley2019semantickitti}
J.~Behley, M.~Garbade, A.~Milioto, J.~Quenzel, S.~Behnke, C.~Stachniss, and J.~Gall, ``Semantickitti: A dataset for semantic scene understanding of lidar sequences,'' in \emph{Proceedings of the IEEE/CVF international conference on computer vision}, 2019, pp. 9297--9307.

\bibitem{achanta2012slic}
R.~Achanta, A.~Shaji, K.~Smith, A.~Lucchi, P.~Fua, and S.~S{\"u}sstrunk, ``Slic superpixels compared to state-of-the-art superpixel methods,'' \emph{IEEE transactions on pattern analysis and machine intelligence}, vol.~34, no.~11, pp. 2274--2282, 2012.

\bibitem{kirillov2023segment}
A.~Kirillov, E.~Mintun, N.~Ravi, H.~Mao, C.~Rolland, L.~Gustafson, T.~Xiao, S.~Whitehead, A.~C. Berg, W.-Y. Lo \emph{et~al.}, ``Segment anything,'' in \emph{Proceedings of the IEEE/CVF International Conference on Computer Vision}, 2023, pp. 4015--4026.

\bibitem{zhuang2021perception}
Z.~Zhuang, R.~Li, K.~Jia, Q.~Wang, Y.~Li, and M.~Tan, ``Perception-aware multi-sensor fusion for 3d lidar semantic segmentation,'' in \emph{Proceedings of the IEEE/CVF International Conference on Computer Vision}, 2021, pp. 16\,280--16\,290.

\bibitem{wu2023cross}
Y.~Wu, M.~Xing, Y.~Zhang, Y.~Xie, J.~Fan, Z.~Shi, and Y.~Qu, ``Cross-modal unsupervised domain adaptation for 3d semantic segmentation via bidirectional fusion-then-distillation,'' in \emph{Proceedings of the 31st ACM International Conference on Multimedia}, 2023, pp. 490--498.

\bibitem{yan20222dpass}
X.~Yan, J.~Gao, C.~Zheng, C.~Zheng, R.~Zhang, S.~Cui, and Z.~Li, ``2dpass: 2d priors assisted semantic segmentation on lidar point clouds,'' in \emph{European Conference on Computer Vision}.\hskip 1em plus 0.5em minus 0.4em\relax Springer, 2022, pp. 677--695.

\bibitem{dai2017scannet}
A.~Dai, A.~X. Chang, M.~Savva, M.~Halber, T.~Funkhouser, and M.~Nie{\ss}ner, ``Scannet: Richly-annotated 3d reconstructions of indoor scenes,'' in \emph{Proceedings of the IEEE conference on computer vision and pattern recognition}, 2017, pp. 5828--5839.

\bibitem{armeni2017joint}
I.~Armeni, S.~Sax, A.~R. Zamir, and S.~Savarese, ``Joint 2d-3d-semantic data for indoor scene understanding,'' \emph{arXiv preprint arXiv:1702.01105}, 2017.

\bibitem{wu2022dual}
Z.~Wu, Y.~Wu, G.~Lin, J.~Cai, and C.~Qian, ``Dual adaptive transformations for weakly supervised point cloud segmentation,'' in \emph{European conference on computer vision}.\hskip 1em plus 0.5em minus 0.4em\relax Springer, 2022, pp. 78--96.

\bibitem{pan2024less}
Z.~Pan, N.~Zhang, W.~Gao, S.~Liu, and G.~Li, ``Less is more: Label recommendation for weakly supervised point cloud semantic segmentation,'' in \emph{Proceedings of the AAAI Conference on Artificial Intelligence}, vol.~38, no.~5, 2024, pp. 4397--4405.

\bibitem{zhang2021perturbed}
Y.~Zhang, Y.~Qu, Y.~Xie, Z.~Li, S.~Zheng, and C.~Li, ``Perturbed self-distillation: Weakly supervised large-scale point cloud semantic segmentation,'' in \emph{Proceedings of the IEEE/CVF international conference on computer vision}, 2021, pp. 15\,520--15\,528.

\bibitem{unal2022scribble}
O.~Unal, D.~Dai, and L.~Van~Gool, ``Scribble-supervised lidar semantic segmentation,'' in \emph{Proceedings of the IEEE/CVF Conference on Computer Vision and Pattern Recognition}, 2022, pp. 2697--2707.

\bibitem{genova2021learning}
K.~Genova, X.~Yin, A.~Kundu, C.~Pantofaru, F.~Cole, A.~Sud, B.~Brewington, B.~Shucker, and T.~Funkhouser, ``Learning 3d semantic segmentation with only 2d image supervision,'' in \emph{2021 International Conference on 3D Vision (3DV)}.\hskip 1em plus 0.5em minus 0.4em\relax IEEE, 2021, pp. 361--372.

\bibitem{liang2019multi}
M.~Liang, B.~Yang, Y.~Chen, R.~Hu, and R.~Urtasun, ``Multi-task multi-sensor fusion for 3d object detection,'' in \emph{Proceedings of the IEEE/CVF Conference on Computer Vision and Pattern Recognition}, 2019, pp. 7345--7353.

\bibitem{liang2018deep}
M.~Liang, B.~Yang, S.~Wang, and R.~Urtasun, ``Deep continuous fusion for multi-sensor 3d object detection,'' in \emph{Proceedings of the European conference on computer vision (ECCV)}, 2018, pp. 641--656.

\bibitem{jaritz2020xmuda}
M.~Jaritz, T.-H. Vu, R.~d. Charette, E.~Wirbel, and P.~P{\'e}rez, ``xmuda: Cross-modal unsupervised domain adaptation for 3d semantic segmentation,'' in \emph{Proceedings of the IEEE/CVF conference on computer vision and pattern recognition}, 2020, pp. 12\,605--12\,614.

\bibitem{10330760}
A.~Xiao, D.~Guan, X.~Zhang, and S.~Lu, ``Domain adaptive lidar point cloud segmentation with 3d spatial consistency,'' \emph{IEEE Transactions on Multimedia}, vol.~26, pp. 5536--5547, 2024.

\bibitem{9913730}
T.~Weng, J.~Xiao, F.~Yan, and H.~Jiang, ``Context-aware 3d point cloud semantic segmentation with plane guidance,'' \emph{IEEE Transactions on Multimedia}, vol.~25, pp. 6653--6664, 2023.

\bibitem{wang2023survey}
J.~Wang, Y.~Liu, H.~Tan, and M.~Zhang, ``A survey on weakly supervised 3d point cloud semantic segmentation,'' \emph{IET Computer Vision}, 2023.

\bibitem{zhou2016learning}
B.~Zhou, A.~Khosla, A.~Lapedriza, A.~Oliva, and A.~Torralba, ``Learning deep features for discriminative localization,'' in \emph{Proceedings of the IEEE conference on computer vision and pattern recognition}, 2016, pp. 2921--2929.

\bibitem{li2022hybridcr}
M.~Li, Y.~Xie, Y.~Shen, B.~Ke, R.~Qiao, B.~Ren, S.~Lin, and L.~Ma, ``Hybridcr: Weakly-supervised 3d point cloud semantic segmentation via hybrid contrastive regularization,'' in \emph{Proceedings of the IEEE/CVF conference on computer vision and pattern recognition}, 2022, pp. 14\,930--14\,939.

\bibitem{xu2022gmflow}
H.~Xu, J.~Zhang, J.~Cai, H.~Rezatofighi, and D.~Tao, ``Gmflow: Learning optical flow via global matching,'' in \emph{Proceedings of the IEEE/CVF conference on computer vision and pattern recognition}, 2022, pp. 8121--8130.

\bibitem{caesar2020nuscenes}
H.~Caesar, V.~Bankiti, A.~H. Lang, S.~Vora, V.~E. Liong, Q.~Xu, A.~Krishnan, Y.~Pan, G.~Baldan, and O.~Beijbom, ``nuscenes: A multimodal dataset for autonomous driving,'' in \emph{Proceedings of the IEEE/CVF conference on computer vision and pattern recognition}, 2020, pp. 11\,621--11\,631.

\bibitem{Butler:ECCV:2012}
D.~J. Butler, J.~Wulff, G.~B. Stanley, and M.~J. Black, ``A naturalistic open source movie for optical flow evaluation,'' in \emph{European Conf. on Computer Vision (ECCV)}, ser. Part IV, LNCS 7577, {A. Fitzgibbon et al. (Eds.)}, Ed.\hskip 1em plus 0.5em minus 0.4em\relax Springer-Verlag, Oct. 2012, pp. 611--625.

\bibitem{he2016deep}
K.~He, X.~Zhang, S.~Ren, and J.~Sun, ``Deep residual learning for image recognition,'' in \emph{Proceedings of the IEEE conference on computer vision and pattern recognition}, 2016, pp. 770--778.

\bibitem{ronneberger2015u}
O.~Ronneberger, P.~Fischer, and T.~Brox, ``U-net: Convolutional networks for biomedical image segmentation,'' in \emph{Medical Image Computing and Computer-Assisted Intervention--MICCAI 2015: 18th International Conference, Munich, Germany, October 5-9, 2015, Proceedings, Part III 18}.\hskip 1em plus 0.5em minus 0.4em\relax Springer, 2015, pp. 234--241.

\bibitem{tang2020searching}
H.~Tang, Z.~Liu, S.~Zhao, Y.~Lin, J.~Lin, H.~Wang, and S.~Han, ``Searching efficient 3d architectures with sparse point-voxel convolution,'' in \emph{Computer Vision--ECCV 2020: 16th European Conference, Glasgow, UK, August 23--28, 2020, Proceedings, Part XXVIII}.\hskip 1em plus 0.5em minus 0.4em\relax Springer, 2020, pp. 685--702.

\bibitem{vaswani2017attention}
A.~Vaswani, N.~Shazeer, N.~Parmar, J.~Uszkoreit, L.~Jones, A.~N. Gomez, {\L}.~Kaiser, and I.~Polosukhin, ``Attention is all you need,'' \emph{Advances in neural information processing systems}, vol.~30, 2017.

\bibitem{renyi1961measures}
A.~R{\'e}nyi, ``On measures of entropy and information,'' in \emph{Proceedings of the Fourth Berkeley Symposium on Mathematical Statistics and Probability, Volume 1: Contributions to the Theory of Statistics}, vol.~4.\hskip 1em plus 0.5em minus 0.4em\relax University of California Press, 1961, pp. 547--562.

\bibitem{hinton2015distilling}
G.~Hinton, O.~Vinyals, and J.~Dean, ``Distilling the knowledge in a neural network,'' \emph{arXiv preprint arXiv:1503.02531}, 2015.

\bibitem{berman2018lovasz}
M.~Berman, A.~R. Triki, and M.~B. Blaschko, ``The lov{\'a}sz-softmax loss: A tractable surrogate for the optimization of the intersection-over-union measure in neural networks,'' in \emph{Proceedings of the IEEE conference on computer vision and pattern recognition}, 2018, pp. 4413--4421.

\bibitem{geiger2013vision}
A.~Geiger, P.~Lenz, C.~Stiller, and R.~Urtasun, ``Vision meets robotics: The kitti dataset,'' \emph{The International Journal of Robotics Research}, vol.~32, no.~11, pp. 1231--1237, 2013.

\bibitem{qi2017pointnet}
C.~R. Qi, H.~Su, K.~Mo, and L.~J. Guibas, ``Pointnet: Deep learning on point sets for 3d classification and segmentation,'' in \emph{Proceedings of the IEEE conference on computer vision and pattern recognition}, 2017, pp. 652--660.

\bibitem{zhang2020polarnet}
Y.~Zhang, Z.~Zhou, P.~David, X.~Yue, Z.~Xi, B.~Gong, and H.~Foroosh, ``Polarnet: An improved grid representation for online lidar point clouds semantic segmentation,'' in \emph{Proceedings of the IEEE/CVF Conference on Computer Vision and Pattern Recognition}, 2020, pp. 9601--9610.

\bibitem{cortinhal2020salsanext}
T.~Cortinhal, G.~Tzelepis, and E.~Erdal~Aksoy, ``Salsanext: Fast, uncertainty-aware semantic segmentation of lidar point clouds,'' in \emph{Advances in Visual Computing: 15th International Symposium, ISVC 2020, San Diego, CA, USA, October 5--7, 2020, Proceedings, Part II 15}.\hskip 1em plus 0.5em minus 0.4em\relax Springer, 2020, pp. 207--222.

\bibitem{zhu2021cylindrical}
X.~Zhu, H.~Zhou, T.~Wang, F.~Hong, Y.~Ma, W.~Li, H.~Li, and D.~Lin, ``Cylindrical and asymmetrical 3d convolution networks for lidar segmentation,'' in \emph{Proceedings of the IEEE/CVF conference on computer vision and pattern recognition}, 2021, pp. 9939--9948.

\bibitem{liong2020amvnet}
V.~E. Liong, T.~N.~T. Nguyen, S.~Widjaja, D.~Sharma, and Z.~J. Chong, ``Amvnet: Assertion-based multi-view fusion network for lidar semantic segmentation,'' \emph{arXiv preprint arXiv:2012.04934}, 2020.

\bibitem{xu2021rpvnet}
J.~Xu, R.~Zhang, J.~Dou, Y.~Zhu, J.~Sun, and S.~Pu, ``Rpvnet: A deep and efficient range-point-voxel fusion network for lidar point cloud segmentation,'' in \emph{Proceedings of the IEEE/CVF International Conference on Computer Vision}, 2021, pp. 16\,024--16\,033.

\bibitem{ando2023rangevit}
A.~Ando, S.~Gidaris, A.~Bursuc, G.~Puy, A.~Boulch, and R.~Marlet, ``Rangevit: Towards vision transformers for 3d semantic segmentation in autonomous driving,'' in \emph{Proceedings of the IEEE/CVF Conference on Computer Vision and Pattern Recognition}, 2023, pp. 5240--5250.

\bibitem{lai2023spherical}
X.~Lai, Y.~Chen, F.~Lu, J.~Liu, and J.~Jia, ``Spherical transformer for lidar-based 3d recognition,'' in \emph{Proceedings of the IEEE/CVF Conference on Computer Vision and Pattern Recognition}, 2023, pp. 17\,545--17\,555.

\bibitem{10128757}
L.~Zhao, H.~Zhou, X.~Zhu, X.~Song, H.~Li, and W.~Tao, ``Lif-seg: Lidar and camera image fusion for 3d lidar semantic segmentation,'' \emph{IEEE Transactions on Multimedia}, vol.~26, pp. 1158--1168, 2024.

\bibitem{wu2021redal}
T.-H. Wu, Y.-C. Liu, Y.-K. Huang, H.-Y. Lee, H.-T. Su, P.-C. Huang, and W.~H. Hsu, ``Redal: Region-based and diversity-aware active learning for point cloud semantic segmentation,'' in \emph{Proceedings of the IEEE/CVF international conference on computer vision}, 2021, pp. 15\,510--15\,519.

\bibitem{huang2024opoca}
W.~Huang, P.~Zou, Y.~Xia, C.~Wen, Y.~Zang, C.~Wang, and G.~Zhou, ``Opoca: One point one class annotation for lidar point cloud semantic segmentation,'' \emph{IEEE Transactions on Geoscience and Remote Sensing}, 2024.

\bibitem{choy20194d}
C.~Choy, J.~Gwak, and S.~Savarese, ``4d spatio-temporal convnets: Minkowski convolutional neural networks,'' in \emph{Proceedings of the IEEE/CVF conference on computer vision and pattern recognition}, 2019, pp. 3075--3084.

\end{thebibliography}

\ifCLASSOPTIONcaptionsoff
  \newpage
\fi

\end{document}